\documentclass[letterpaper, 10 pt, conference]{ieeeconf}  

\IEEEoverridecommandlockouts                              

\usepackage{xcolor}
\usepackage{amsmath,amsfonts,amssymb}
\usepackage{array,multirow,graphicx}
\usepackage{subfigure}

\usepackage{xspace}  
\newcommand{\tbc}{BC\xspace}
\newcommand{\tbcwmse}{BC wMSE\xspace}
\newcommand{\tbcwMSE}{BC wMSE + Orientation\xspace}
\newcommand{\tdiffsim}{DiffSim\xspace}
\newcommand{\tdiff}{Diff.\xspace}
\newcommand{\tmgail}{MGAIL\xspace}
\newcommand{\tcollision}{Collision\xspace}
\newcommand{\tcol}{Col.\xspace}
\newcommand{\tbcgaussll}{BC Gaussian-LL\xspace}
\newcommand{\tidm}{IDM\xspace}
\newcommand{\tbcgmmll}{BC GMM-LL\xspace}
\newcommand{\tbcll}{BC-LL\xspace}
\newcommand{\tmse}{MSE\xspace}
\newcommand{\twmse}{wMSE\xspace}
\newcommand{\torientation}{Orientation\xspace}
\newcommand{\tdiffsimmse}{\tdiffsim \tmse}
\newcommand{\tdiffsimwmse}{\tdiffsim \twmse}
\newcommand{\tmgaildiffsimwmse}{\tmgail+ \tdiffsim \twmse}
\newcommand{\tdiffsimwmsecollision}{\tdiffsim \twmse + \tcollision}
\newcommand{\tmgaildiffsimwmsecollision}{\tmgail + \tdiffsim \twmse + \tcollision}
\newcommand{\tmgailbcll}{\tmgail + \tbcll}
\newcommand{\tmgaildiffwmsecol}{\tmgail + \tdiff \twmse + \tcol}

\newcommand{\ttbc}{\texttt{\tbc}\xspace}
\newcommand{\ttorientation}{\texttt{\torientation}\xspace}
\newcommand{\ttbcwmse}{\texttt{\tbcwmse}\xspace}
\newcommand{\ttbcwMSE}{\texttt{\tbcwMSE}\xspace}
\newcommand{\ttdiffsim}{\texttt{\tdiffsim}\xspace}

\newcommand{\ttbcgaussll}{\texttt{\tbcgaussll}\xspace}
\newcommand{\ttbcgmmll}{\texttt{\tbcgmmll}\xspace}
\newcommand{\ttbcll}{\texttt{\tbcll}\xspace}

\newcommand{\ttdiffsimmse}{\texttt{\tdiffsimmse}\xspace}
\newcommand{\ttdiffsimwmse}{\texttt{\tdiffsimwmse}\xspace}
\newcommand{\ttmgaildiffsimwmse}{\texttt{\tmgaildiffsimwmse}\xspace}
\newcommand{\ttdiffsimwmsecollision}{\texttt{\tdiffsimwmsecollision}\xspace}
\newcommand{\ttmgaildiffsimwmsecollision}{\texttt{\tmgaildiffsimwmsecollision}\xspace}
\newcommand{\ttmgailbcll}{\texttt{\tmgailbcll}\xspace}

\title{\LARGE \bf
Analyzing Closed-loop Training Techniques for Realistic Traffic Agent Models in Autonomous Highway Driving Simulations
}

\author{Matthias Bitzer$^{1}$, Reinis Cimurs$^{2}$, Benjamin Coors$^{2}$, Johannes Goth$^{1}$, Sebastian Ziesche$^{1}$,\\ 
Philipp Geiger$^{1}$, Maximilian Naumann$^{1}$
\thanks{$^{1}$Matthias Bitzer, Johannes Goth, Sebastian Ziesche, Philipp Geiger and Maximilian Naumann are with Bosch Center for Artificial Intelligence (BCAI), Robert-Bosch-Campus 1,
71272 Renningen, Germany
        {\tt\small \{matthias.bitzer, johannes.goth, sebastian.ziesche, philipp.geiger, maximillian.naumann\}@de.bosch.com}}%
\thanks{$^{2}$ Reinis Cimurs and Benjamin Coors are with Robert Bosch GmbH, Crossdomain Computing division, Wienerstrasse 42-46, 70469 Stuttgart-Feuerbach, Germany
        {\tt\small \{reinis.cimurs, benjamin.coors\}@de.bosch.com}}%
}

\begin{document}

\maketitle
\thispagestyle{empty}
\pagestyle{empty}

\begin{abstract}

Simulation plays a crucial role in the rapid development and safe deployment of autonomous vehicles. Realistic traffic agent models are indispensable for bridging the gap between simulation and the real world. Many existing approaches for imitating human behavior are based on learning from demonstration. However, these approaches are often constrained by focusing on individual training strategies. Therefore, to foster a broader understanding of realistic traffic agent modeling,
in this paper, we provide an extensive comparative analysis of different training principles, with a focus on closed-loop methods for highway driving simulation. We experimentally compare (i) open-loop vs.\ closed-loop multi-agent training,  (ii) adversarial vs.\ deterministic supervised training, (iii) the impact of reinforcement losses, and (iv) the impact of training alongside log-replayed agents to identify suitable training techniques for realistic agent modeling. Furthermore, we identify promising combinations of different closed-loop training methods. 

\end{abstract}

\section{INTRODUCTION}

Modeling the behavior of traffic participants is a crucial component in the development process of autonomous driving systems. 
Multi-agent driver models are utilized, for example, in simulation \cite{suo2021trafficsim, bergamini2021simnet} to benchmark planners or in planning systems themselves to reason about the behavior of other traffic participants \cite{driggs2017integrating}. 
However, most deployed driver models are rule-based and are not able to capture behavior outside their manually specified rules. 
Data-driven driver models offer an alternative that is able to learn behavior directly from real-world data.

\begin{figure*}[h]
    \centering
    \includegraphics[width=1\linewidth]{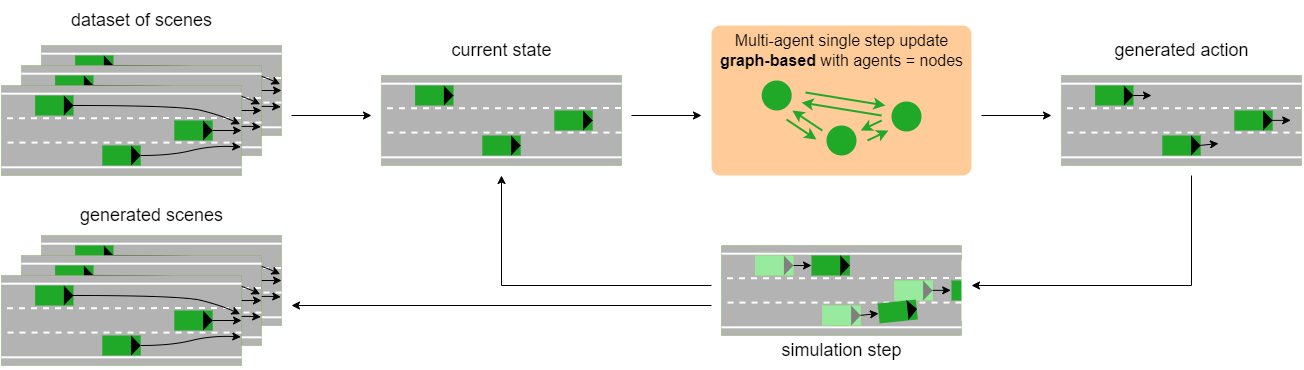}
    \caption{Our problem setting - closed-loop multi-agent policy learning for autonomous driving.}
    \label{fig:diffsim_overview}
\end{figure*}

Many general learning methods have been proposed to learn multi-agent driving policies, including simple one-step supervised learning \cite{feng2023trafficgen, bergamini2021simnet}, closed-loop deterministic imitation learning \cite{scheel2021urban,pmlr-v205-karkus23a, Suo_2023_CVPR}, 
adversarial imitation learning \cite{kuefler2017imitating, tabatabaie2023interaction,igl2022symphony} and combinations of imitation with Reinforcement Learning (RL) \cite{scheel2021urban, scibior2021itra, lioutas2022titrated, zhang2023learning, xu2023bits,lu2023imitation}.
Most recent  methods have in common that the training is executed closed-loop, meaning that the model directly executes a sequence of actions with a differentiable forward model in the loop (see Figure \ref{fig:diffsim_overview}), instead of predicting the next action given a ground-truth state (which we refer to as open-loop).
This enables the policy to reason about the future consequences of a given action. While many of the aforementioned works propose new training methods and/or architectures, a systematic, comparative, empirical study of high-level training principles is often missing. But such a study is important to understand the sim-to-real gap induced by different training paradigms, in particular when using them for safety testing.
Our goal in this paper is to systematically compare and analyze different training paradigms for multi-agent driver models, with a particular focus on closed-loop methods in highway scenarios.
The analysis includes the following dimensions:
\begin{itemize}
    \item \textit{Closed-loop vs. open-loop:} While it is already evident that closed-loop training is beneficial, we reevaluate this claim over a larger set of training methods. 
    \item \textit{Deterministic supervised learning vs. probabilistic adversarial learning:} Recent methods used Model-based Generative Adversarial Imitation Learning (MGAIL) \cite{bronstein2022hierarchical} to train a probabilistic driver model. However, also training purely deterministically in a closed-loop supervised fashion is possible \cite{scheel2021urban,pmlr-v205-karkus23a, Suo_2023_CVPR}. It is unclear how both compare.
    \item \textit{Pure Imitation vs. additional reinforcement learning signal:} There is some evidence that training a policy with imitation combined with reinforcement learning is beneficial \cite{lu2023imitation}. Thus, we include a reinforcement signal in our analysis of closed-loop trainings. 
    \item \textit{Log-replay vs. multi-agent training:} Most methods either propose a single-agent training alongside log-replayed agents or a multi-agent learning scheme. The comparison of the two schemes is infrequent.
\end{itemize}
Each training method has its theoretical benefits and shortcomings. 
Policies trained via deterministic supervised training might lack diverse behavior, but can be trained in a stable way. 
Adversarial training might, in theory, be able to match state distributions perfectly but are hard to train and it is unclear if the discriminator matches the distribution on features that are actually relevant for the driving task. 
Furthermore, one can enforce driving properties that are important for the driving task, such as collision avoidance, via a reinforcement learning signal but it is unclear how much this impacts other realism properties.

To execute our study, we propose an intuitive multi-agent policy parameterization that can be employed in all training methods with minimal adjustments. 
The architecture is based on a multi-agent Graph Neural Net (GNN) encoder that is invariant to rotation and translation of the scene and only operates on differentiable features. 
The decoder can be a stochastic head (for MGAIL), a deterministic head (for deterministic supervised learning) or a head that maps to discriminator scores.
Our experiments are executed on the exiD dataset \cite{moers2022exid} consisting of sixteen hours of real-world driving data that focuses on agent-to-agent interaction on highway on-ramp and off-ramp scenarios.
We identify the following useful strategies for training realistic traffic agent models on these scenarios:
\begin{enumerate}
    \item 
    All investigated closed-loop training strategies show superior performance over their open-loop counterparts.
    \item A reinforcement signal improves crucial metrics such as collision rate, but easily deteriorates other realism metrics.
    \item Deterministic supervised learning can be competitive with probabilistic adversarial learning in highway scenarios.
    \item Combining different closed-loop policy learning strategies improves crucial metrics such as collision rate while maintaining realism.
\end{enumerate}

\section{Related Work}
\label{sec:related work}

\textbf{Rule-based driver models.} The driver modeling task is often solved by employing rule-based methods for creating a decision-making policy.
Generally, rule-based driver models express the driving task as a set of parameterized functions,
such as the Intelligent Driver Model (\tidm) \cite{treiber2000congested} and its extensions \cite{kesting2010enhanced, zhou2016impact, sharath2020enhanced}. 
Their popularity stems from the simple implementation and parameterization based on ego agent's velocity, distance to other vehicles, and the velocity difference.
To obtain the best performing parameters for this method, data driven approaches have been used in \cite{kanagaraj2013evaluation, pourabdollah2017calibration, zhang2021comprehensive}.
Other works \cite{wiedemann1974simulation, chakroborty1999evaluation, krauss1997metastable, gipps1981behavioural, kesting2007general} parameterize car following models based on surrounding vehicle features. While rule-based models are highly interpretable and computationally efficient, they lack expressiveness w.r.t. realism of the driver behavior and suffer from poor generalization \cite{teng2023motion, chen2024data}. To address these issues, different learning-based methods have come to the forefront in autonomous driving research.

\textbf{Driver modeling with Reinforcement Learning.} In RL implementations \cite{shiroshita2020behaviorally, Chen_2021_ICCV, Toromanoff_2020_CVPR} the driving policy is trained to maximize a pre-specified reward obtained through the interaction with the environment. Typically, the environment dynamics are non-differentiable or even unknown. Designing a reward function that captures realistic motion behavior is extremely difficult due to the subtle intricacies of human decision making.
Assuming that the reward is a function purely defined by the state and is dynamics invariant \cite{fu2017learning}, its function can be inferred through Inverse Reinforcement Learning (IRL) \cite{zheng2021objective}.
Yet, IRL is expensive to run and difficult to scale \cite{wu2020efficient}.

\textbf{Driver Modeling with Imitation Learning.} Imitation Learning (IL) is used to train a policy solely from expert demonstrations. Classically, Behavioral Cloning (BC) can be used in an open-loop setting to obtain a trained policy directly from ground truth demonstrations \cite{feng2023trafficgen, bergamini2021simnet, bansal2018chauffeurnet}.
However, these methods suffer from distribution shift in long tail rollouts \cite{ross2011reduction}.
This can be alleviated with applying augmentations \cite{bergamini2021simnet} or training in closed-loop where the training is performed directly in a sequential decision-making manner.
Here, differential simulation accumulates the loss over multiple steps \cite{suo2021trafficsim, pmlr-v205-karkus23a, scibior2021itra, igl2022symphony, Suo_2023_CVPR}.
This requires either augmenting the input data if working with rasterized representations of environment \cite{scibior2021itra} or using vector representations \cite{gao2020vectornet} to have fully differentiable features.
However, pure IL can be insufficient to learn safe and reliable policies due to a rareness of critical scenarios in the ground truth data \cite{lu2023imitation}.
Therefore, loss in IL is augmented with RL rewards or common sense losses.
In \cite{scheel2021urban, scibior2021itra, lioutas2022titrated, zhang2023learning, xu2023bits} a level of infraction is used to penalize the collisions and road departures in the loss.

\textbf{Driver Modeling with Generative Adversarial Learning.} Adversarial learning is an alternative approach for imitating expert behavior. 
Here, the loss function is replaced with a learned discriminator, and the driving policy is trained to fool the discriminator. Foundational work that combined adversarial and policy learning was done with Generative Adversarial IL (GAIL) \cite{kuefler2017imitating, tabatabaie2023interaction,song2016mgail,bronstein2022hierarchical}. These methods typically need an interplay between discriminator optimization and solving the RL problem. Importantly, when the environment dynamics is known and differentiable, GAIL can be replaced with model-based generative adversarial IL (MGAIL), which can be used for closed-loop training \cite{song2016mgail}. 
In the light of driver modeling, it was used for single-agent planner learning \cite{bronstein2022hierarchical} and multi-agent driver-modeling for simulations \cite{igl2022symphony}. 
While those methods alleviate the necessity to specify a loss, they are difficult to train \cite{pmlr-v80-mescheder18a, dadashi2020primal} and can still suffer from a lack of realism in feature distribution if not addressed in the training process \cite{yan2023learning}.

\section{Problem Formulation and Base Policy}
\label{sec:method}
In this section, we phrase our problem formulation and introduce our base policy parameterization. In Section \ref{sec:training} we outline the different training paradigms for analysis.

\textbf{Problem Formulation.} 
Our driver modeling goal is to learn a multi-agent policy $\pi_{\theta}(\mathbf{a}_{t}|\mathbf{s}_{t})$,
where $\mathbf{a_{t}}=(a_{1,t},\dots,a_{M,t})$
denotes the actions of $M$ traffic participants at time step $t$ and $\mathbf{s_{t}}$ denotes the states of all agents, which contain information like the position, speed etc.\ , and the local map, at time step $t$. 
Given a set of ground truth trajectories $\mathcal{D}=\{\tau_{i}\}_{i=1}^{N}$ with $\tau_{i}=(\mathbf{s}_{0}^{(i)},\mathbf{a}_{1}^{(i)},\dots,\mathbf{s}_{T}^{(i)},\mathbf{a}_{T}^{(i)})$ the goal is to learn the parameters $\theta$ of the policy such that it induces a distribution over trajectories 
$p(\tau)=p(\mathbf{s}_{0})\prod_{t=0}^{T}p(\mathbf{s}_{t+1}|\mathbf{a}_{t},\mathbf{s}_{t})\pi_{\theta}(\mathbf{a}_{t}|\mathbf{s}_{t})$
that resembles the distribution of trajectories in $\mathcal{D}$. Here, we assume a known, deterministic and differentiable transition model $p(\mathbf{s}_{t+1}|\mathbf{a}_{t},\mathbf{s}_{t})$, which is a common assumption in driver modeling \cite{scheel2021urban,Suo_2023_CVPR}. 
In particular, we use position delta actions for each agent (detailed in differential update step).
The main challenge of driver modeling consists of modeling the underlying policy $\pi_{\theta}$ and choosing the learning method to fit the generated trajectories to the ground truth trajectories. 
In the following paragraphs we propose our policy parameterization, that induces an intuitive inductive bias for multi-agent driving. 

\textbf{Policy architecture.} Our multi-agent policy $\pi_{\theta}(\mathbf{a}_{t}|\mathbf{s}_{t})$ follows an encoder-decoder architecture. The encoder takes in the multi-agent state $\mathbf{s}_{t}$ and returns an encoding for each agent:
\begin{align}
    \mathbf{h}_{t}:=[h_{1,t},\dots,h_{M,t}] = \mathrm{Encoder}(\mathbf{s}_{t})
\end{align}
For multi-agent scenarios it is crucial that the $\mathrm{Encoder}$ can deal with different number of agents in the scene. Here, it is natural to define a graph over agents and use a GNN.
The agents are the nodes in a locally connected graph. 
For a given multi-agent state $\mathbf{s}_{t}$ we extract initial node features $\mathbf{n}^{(0)}=[n^{(0)}_{1},\dots,n^{(0)}_{M}]$ and initial edge features $\mathbf{e}^{(0)}=[e^{(0)}_{i\xrightarrow{}j}]_{i\neq j}$ and process them via repeating message-passing layers with index $k=0,\dots,K-1$
\begin{align}
    \mathbf{n}^{(k+1)},\mathbf{e}^{(k+1)}=\mathrm{MessagePassing}(\mathbf{n}^{(k)}, \mathbf{e}^{(k)})
\end{align}
The output of the encoder is given via $\mathbf{h}_{t}=\mathbf{n}^{(K)}$. We elaborate on the details of the initial node and edge features as well as the message passing algorithm in the paragraphs below. 

Depending on the training method (see Section \ref{sec:training}) the decoder of the policy either maps the agent-wise encoding directly to the 2D action space via a weight-shared MLP,
\begin{align}
    \mathbf{a}_{t}=[\mathrm{MLP}(h_{1,t}),\dots,\mathrm{MLP}(h_{M,t})]=\mathrm{Decoder}(\mathbf{h}_{t}).
\end{align}
or each MLP maps to the parameters $\theta_{i,t}$ of a 2D distribution like a Gaussian or Gaussian Mixture (GMM) which defines an agent-specific distribution over our action space. 


\textbf{Node features.} The initial node features of the GNN contain information about the agents' poses and kinematics, the agents' local map, and further information about the agent, like the agent type and its dimensions.
First, we embed poses, kinematics, agent type and dimensions into a single embedding vector $h_{i,t}$ via an MLP. 
The map is represented as a sequence of line strings in the local coordinate system of the agent. Those line strings are associated with the boundary lines of each lane segment. 
The line strings are embedded with Multi-Head Attention (MHA) resulting in a sequence of line embeddings $[l_{i,j}]_{j=1}^{J}$. 
The initial embedding to the GNN is computed by fusing map information via cross attention (with residual) from $h_{i,t}$ to $[l_{i,j}]_{j=1}^{J}$.
All features are computed in the local reference frame of each agent and are thus rotation and translation invariant.


\textbf{Edge features.} 
The edge features capture relevant properties for interaction between source and target agents. 
The edge features therefore consist of the distance and velocity difference between source and target agents, relative position history between source and target agent for $t-2:t$ in target agents' coordinate frame, heading difference\footnote{Using $\cos$ and $\sin$ of the difference, to ensure continuity of the feature.}, and the time to collision between the source and target agent, clipped at a maximum value of 10~s. 
All these features are invariant to rotations and translations of the agent pair.

\textbf{Message-Passing.} Intuitively, our GNN has the inductive bias that the agents are the nodes in the graph, their interaction is modeled via edges, and the reasoning over others and the resulting behavior is computed with the message-passing steps, as principally proposed by \cite{casas20spagnn}. 
Concretely, our message passing module employs an edge model that updates the edge features via $e^{(k+1)}_{i\xrightarrow{}j}=\mathrm{MLP}([h_{i}^{(k)},h_{j}^{(k)},e^{(k)}_{i\xrightarrow{}j}])$ and uses cross-attention (with residual) between $h_{j}^{(k)}$ and $[e^{(k+1)}_{i\xrightarrow{}j}]_{i=1}^{M}$ to get $h_{j}^{(k+1)}$. In this way, each target agent can focus on the relevant agents that might interact with it.

\textbf{Differentiable update step.} A crucial component to enabling closed-loop training is a differentiable forward step, that enables propagating gradients through the steps of the trajectory. 
As actions $a_{i,t}$, we use the position deltas in the local reference frame of each agent, i.e.,
for $i$'s local 2D position at time $t+1$, we have $(x_{i,t+1},y_{i,t+1})=(x_{i,t},y_{i,t})+a_{i,t}$. 
After transforming this to the global reference frame, the position and the heading of each agent are updated (see Figure \ref{fig:diffsim_overview}). 
The new multi-agent state $\mathbf{s}_{t+1}$ is calculated via differentiable transformations of all features given the new locations and headings. 
Crucially, this enables MGAIL and differentiable simulation training to propagate future error through time to earlier actions. 
We investigate the impact of different variants of closed-loop training in our experiments.

\section{Compared Training Approaches}
\label{sec:training}
We introduce the different training paradigms we analyse for learning realistic driver models.

\textbf{Behavioral Cloning.} The first intuitive training method is supervised one-step imitation learning, also called behavioral cloning. 
Here, we minimize an imitation loss over the one-step state-action distribution. For example, one might minimize the expected negative log likelihood of the policy under the one-step data distribution,
$
    \mathcal{L}_{\mathrm{BC}}  (\theta)=\mathbb{E}_{\mathbf{s},\mathbf{a}\sim\mathcal{D}}[-\mathrm{log}\ \pi_{\theta}(\mathbf{a}|\mathbf{s})].
$
While it appears to be an intuitive principle to train a policy, it has been shown repeatedly \cite{ross2011reduction,ho2016generative} that it can easily lead to compounding errors and unrealistic distributions over (multi-step) trajectories $p(\tau)$. 
However, we use this method for comparison purposes, as pretraining (see Section \ref{sec:result}) and as regularizer (see MGAIL). 

\textbf{Differentiable Simulation.} Since our forward model is differentiable, we can alleviate the compounding error problem via training $\pi_{\theta}$ through differentiable simulation, aka propagating gradients through time. 
Here, we consider the policy to be a deterministic mapping from states to actions $\mathbf{a}_{t}=\pi_{\theta}(\mathbf{s}_{t})$ rather than a probability distribution. 
Given some initial state $\mathbf{s}_{0}$ and a generated trajectory $\tilde{\tau}=(\mathbf{s}_{0},\tilde{\mathbf{a}}_{0}, \tilde{\mathbf{s}}_{1},\dots, \tilde{\mathbf{s}}_{1})$, we use as loss
$
    \mathcal{L}_{\mathrm{DS}}(\theta)=\mathbb{E}_{\mathbf{s}_{0}\sim\mathcal{D}}[\sum_{t=1}^{T}d(\mathbf{s}_{t},\tilde{\mathbf{s}}_{t})|\theta],
$
where $d(\mathbf{s}_{t},\tilde{\mathbf{s}}_{t})$ is a weighted mean squared error (MSE) loss between the $(x,y)$ positions in the states averaged over all agents.
When training via differentiable simulation we pretrain the weights with BC, where we also replace the log likelihood loss with the weighted MSE loss. 

\textbf{MGAIL.} Recent methods \cite{kuefler2017imitating, tabatabaie2023interaction, bronstein2022hierarchical} utilized generative adversarial networks to learn driver models. Here, we train a discriminator $D_{\psi}$ in addition to the policy $\pi_{\theta}$. 
The discriminator is trained to classify states into the ones that come from ground truth and the ones that are generated via the $\pi_{\theta}$. 
It maps from states to probability scores for each agent, thus $D_{\psi}(\mathbf{s})\in [0,1]^{M}$, and is trained via the cross entropy loss
$ 
    \mathcal{L}(\psi)=\mathbb{E}_{\mathbf{s}\sim{\mathcal{D}}}[-\mathrm{log}\, D_{\psi}(\mathbf{s})]+\mathbb{E}_{\mathbf{s}\sim \pi_{\theta}} [-\mathrm{log}\, (1-D_{\psi}(\mathbf{s}))]
$ (here the expectation includes an additional averaging along the time dimension, over all states $\mathbf{s}$ of the individual trajectories).
We parameterize our discriminator in the same way as the policy, except that the $\mathrm{MLP}$ in the decoder maps to $[0,1]$ instead of the 2D action space.

The policy/generator is trained to fool the discriminator and thus minimizes the loss
$
    \mathcal{L}_{\mathrm{MGAIL}}(\theta)=\mathbb{E}_{\mathbf{s}\sim \pi_{\theta}} [\mathrm{log}\, (1-D_{\psi}(\mathbf{s}))].
$
It is important to note that gradients in this loss, can also propagate to previous time points, because of the differentiable forward model. 
Here, the decoder of the policy maps to the parameters of a proper probability distribution and is either parameterized via a Gaussian or Gaussian mixture distribution. 
The loss can be approximated via sampling with the reparameterization trick.

\textbf{Differentiable Collision Loss.} Combining data-based losses with reinforcement learning (RL) losses has been shown to be beneficial for learning planner policies \cite{lu2023imitation}.
Also, for driver modeling, enforcing certain aspects, like preventing collisions, is crucial. 
However, directly applying additional RL-like losses can induce unrealistic trajectories.
To investigate the impact of RL-losses we consider a differentiable collision loss $L_{\mathrm{Collision}}(\theta)$, as proposed in \cite{suo2021trafficsim}, that we use as an auxiliary loss in addition to the data-based losses. 

\textbf{Combination of Methods.} In our experiments, we combine different methods, and denote this by ``+'' signs. 
For example, we can combine differentiable simulation and MGAIL via minimizing the loss
$
    \mathcal{L}(\theta)=\mathcal{L}_{\mathrm{DS}}(\theta)+\mathcal{L}_{\mathrm{MGAIL}}(\theta) 
$
for the generator and train the discriminator as in MGAIL. Similarly, we can combine differentiable simulation with the collision loss, which can be seen as a closed-loop version of the method in \cite{lu2023imitation}. 
Mixing different methods can combine benefits of both methods and alleviate problems that one training principle has when used on its own. 
Importantly, one needs to consider proper weighting of the loss functions.

\section{Experiments \& Results}
\label{sec:result}
In the following, we present our ablation study on the different training methods for modeling highway traffic agents. 
First, we introduce our experimental setup and show the results. In Section \ref{sec:discussion}, we summarize the high-level findings of our experiments.

\textbf{Dataset.} We evaluate on the exiD dataset \cite{moers2022exid}, a real-world trajectory dataset that contains drone-recorded driving data from highway entries and exits in Germany. 
We extract training and evaluation data by cutting the exiD recordings into snippets of 10 second length, which we downsample to a frequency of 2 Hz.
In total, our dataset consists of 5750 recording snippets, which we refer to as rollouts, where all rollouts of one recording are assigned to one split. 
The dataset is organized into a train, validation and test split with 4461, 737 and 552 rollouts, respectively. 
Furthermore, we ensure that each split contains rollouts from each of the seven exiD recording locations. 

\textbf{Simulation setup.} We simulate for the full 10 seconds of our rollouts at 2 Hz. 
In order to enable the computation of the node features, which include the agent's speed and acceleration, we only start controlling an agent after 3 steps of it being present in the rollout. 
This means that an agent effectively performs three initial log-replay steps before it is controlled by the model.

\textbf{Evaluated Methods.}
We compare the methods presented in Section \ref{sec:training} along with their combinations.
Concretely, for open-loop training, we evaluate BC training with maximum likelihood (\ttbcll) using a Gaussian head (\ttbcgaussll) as well as a Gaussian mixture head (\ttbcgmmll). 
We consider training BC with weighted MSE combined with orientation loss (\ttbcwMSE). 
Here, \ttbcwmse refers to an MSE loss where different weights are applied for x and y axis deviations, since lateral motion is predominantly smaller in highway scenarios. 
\ttorientation is a loss expressed as:
\begin{align}
    d_{\textrm{Orientation}}(\hat{\mathbf{a}},\mathbf{a}):=\frac{1}{M}\sum_{i=1}^{M}d_{\textrm{MSE}}((\hat{\delta}_{x,i},\hat{\delta}_{y,i}),(\delta_{x,i},\delta_{y,i})),
\end{align}
where $(\delta_{x,i},\delta_{y,i})$ is the ground-truth normalized heading of the next time step for agent $i$ and $(\hat{\delta}_{x,i},\hat{\delta}_{y,i})$ is the heading that would result from the chosen action. 
$M$ is the number of vehicles.
For closed-loop training, we consider deterministic training through differentiable simulation with a weighted MSE loss (\ttdiffsimwmse) and an unweighted MSE loss (\ttdiffsimmse). 
Furthermore, MGAIL is trained in combination with a BC loss (\ttmgailbcll) and in combination with differential simulation (\ttmgaildiffsimwmse), using a Gaussian or GMM head. 
To investigate the impact of an additional reinforcement loss, we consider adding the differentiable collision loss to differentiable simulation (\ttdiffsimwmsecollision) and to the combination of MGAIL and differentiable simulation (\ttmgaildiffsimwmsecollision).
For all closed-loop methods, we execute a combined multi-agent training and training along log-replay agents.

\begin{table*}
    \centering
    \caption{Loss ablation study for multi-step closed-loop trainings when trained on controlling \textbf{all agents} and evaluated when controlling \textbf{all agents}. }
    \scriptsize
    \begin{tabular}{l|cccccc} 
        \hline
        Method  &
        \begin{tabular}{@{}c@{}} Col. \\ (\%) \end{tabular}  &
        \begin{tabular}{@{}c@{}} Off. \\ (\%) \end{tabular}  &
        \begin{tabular}{@{}c@{}} ADE \\ (m) \end{tabular}  &
        \begin{tabular}{@{}c@{}} Speed \\ JSD $\times 10^{-2}$ \end{tabular}  &
        \begin{tabular}{@{}c@{}} Acc. \\ JSD $\times 10^{-2}$ \end{tabular}  &
        \begin{tabular}{@{}c@{}} $N_{LC}$ \\ JSD $\times 10^{-2}$ \end{tabular}  \\ \hline
        \tidm & 0.8 ± 0.0 & \textbf{0.00 ± 0.0} & 5.31 ± 0.0 & 4.0 ± 0.0 & 17.64 ± 0.0 & 0.23 ± 0.0 \\ \hline
\tbcgaussll                        &                           2.67 ± 0.11 &       6.42 ± 0.42 &      2.22 ± 0.17 &    0.63 ± 0.41 &            8.02 ± 4.6 &                  2.28 ± 0.16 \\
\tbcgmmll                              &                            3.1 ± 0.62 &       7.05 ± 0.78 &      3.42 ± 0.41 &    2.41 ± 0.66 &          23.16 ± 6.52 &                  2.23 ± 0.33 \\        
\tbcwMSE                                    &                            3.1 ± 0.65 &       4.36 ± 0.73 &      4.15 ± 0.34 &     3.51 ± 0.8 &          12.54 ± 4.33 &                   1.5 ± 0.22 \\
\hline
\tdiffsimmse                            &                            1.1 ± 0.74 &       3.05 ± 2.11 &      1.62 ± 0.77 &     0.28 ± 0.3 &           1.82 ± 1.57 &                  0.92 ± 1.03 \\
\tdiffsimwmse                                &                           0.51 ± 0.11 &       0.68 ± 0.31 &      1.25 ± 0.13 &     0.2 ± 0.09 &           1.05 ± 0.38 &                  0.48 ± 0.09 \\
\tdiffsimwmsecollision                   &                           0.39 ± 0.42 &       2.52 ± 1.15 &      3.57 ± 3.31 &    3.24 ± 5.78 &             2.5 ± 1.7 &                   0.66 ± 0.4 \\
\hline
\tmgailbcll (Gauss.)                  &                            1.7 ± 0.58 &        2.95 ± 1.2 &      1.95 ± 0.24 &    0.42 ± 0.18 &           1.82 ± 0.52 &                   0.6 ± 0.56 \\
\tmgailbcll (GMM)                      &                           2.87 ± 2.42 &       2.77 ± 0.52 &       7.8 ± 8.31 &    9.34 ± 10.7 &         18.55 ± 10.69 &                  \textbf{0.19 ± 0.13} \\
\tmgaildiffsimwmse (Gauss.)             &                           0.39 ± 0.06 &       0.59 ± 0.35 &      \textbf{0.93 ± 0.03} &    \textbf{0.08 ± 0.03} &           \textbf{0.86 ± 0.15} &                  0.32 ± 0.15 \\
\tmgaildiffsimwmse (GMM)                  &                           0.35 ± 0.19 &       0.36 ± 0.32 &       1.0 ± 0.18 &     0.11 ± 0.1 &           1.03 ± 0.71 &                  0.31 ± 0.15 \\

\tmgaildiffwmsecol (Gauss.) &                           \textbf{0.17 ± 0.09} &       2.28 ± 0.84 &      1.47 ± 0.09 &    0.23 ± 0.05 &           2.35 ± 0.48 &                  0.26 ± 0.21 \\
\tmgaildiffwmsecol (GMM)      &                           0.17 ± 0.11 &       1.49 ± 0.42 &      1.48 ± 0.16 &    0.26 ± 0.06 &           1.86 ± 0.85 &                  0.35 ± 0.42 \\
\end{tabular}
    \label{tab:train_on_all_eval_on_all}
\end{table*}

\textbf{Training setup.} We train each method for 100 epochs. 
Hyperparameters for each training method are determined with hyperparameter sweeps based on results from evaluation on the validation set. 
Training of BC methods is performed at every step of the rollout, using the method's respective losses.
Differential simulation methods are initialized from a pre-trained \ttbcwMSE model.
Execution is simulated for the full length of the rollout.
MGAIL methods are initialized with weights from a pre-trained \ttbcll model with respective log-likelihood loss.
In single-agent log-replay trainings, we randomly select one controlled agent in the rollout, with a probability proportional to the agent's number of time steps in the scene for each batch element.

\textbf{Evaluation setup.}
All models are evaluated in closed-loop simulation on the test split over all the time steps in the rollout.
In our evaluation, we compare two control settings of the agents.
First, the setting of controlling all agents at once. For the test split of the exiD dataset, this means that we control, on average, about 26 agents per scene.
Since model training is not deterministic, we evaluate their performance variance by training the base BC models with 5 different seed values, then use these model weights to initialize the respective differential simulation and MGAIL methods.
The results are shown in Table \ref{tab:train_on_all_eval_on_all} and Table \ref{tab:train_on_one_eval_on_all} which report the mean and standard deviation of the respective 5 evaluations.
Second, the setting of only controlling one agent, with all other agents being log-replayed.
Here, the controlled agent in the scene is selected deterministically as the agent with the most timesteps in the rollout. 

\textbf{Metrics.}
We employ a variety of metrics to judge a model's quality and realism. Metrics are computed on the test split for agents and frames, where an agent is controlled in a respective frame by the model.

A first set of metrics evaluates the model's infractions. Here, a \textbf{collision rate} is defined as the percentage of agents with at least one collision in a given rollout. 
It is implemented as a polygon intersection check between the bounding boxes of the agents. An \textbf{off-road rate} computes the percentage of frames at which a controlled agent drives off the road, i.e., outside the highway lanes.

We also report the Average Displacement Error (\textbf{ADE}) which is the L2 distance between the ground-truth and generated trajectories for a fixed horizon of 5 seconds. 
This metric measures how closely the generated trajectories follow the ground-truth trajectories.

We evaluate the distributional realism of the models with metrics that compute the Jensen-Shannon Divergence (JSD) between the ground-truth and generated distributions of the agent's \textbf{speed}, \textbf{acceleration} and \textbf{number of lane changes $N_{LC}$}. 
The JSD is computed between histograms with $100$ equisized bins for speed, acceleration and number of lane changes in a range that covers the minimum and maximum of the generated and ground-truth metric values.

\textbf{Results.} Table \ref{tab:train_on_all_eval_on_all} presents our results for all mentioned methods when trained in a combined multi-agent way (non log-replay). 
Here, as baseline we compare the results to \tidm \cite{treiber2000congested} with MOBIL \cite{kesting2007general} to allow for lane changes in highway and on-ramping/off-ramping scenarios.
In Table \ref{tab:train_on_one_eval_on_all}, we show the results of the closed-loop methods when trained alongside log-replay agents. 
Videos of ground-truth and generated scenarios can be found in the supplementary material.

\begin{table*}
    \centering
    \caption{Loss ablation study for multi-step closed-loop trainings when trained on controlling \textbf{one agent} and evaluated when controlling \textbf{all agents}. }
    \scriptsize
    \begin{tabular}{l|cccccc} 
        \hline
        Method  &
        \begin{tabular}{@{}c@{}} Col. \\ (\%) \end{tabular}  &
        \begin{tabular}{@{}c@{}} Off. \\ (\%) \end{tabular}  &
        \begin{tabular}{@{}c@{}} ADE \\ (m) \end{tabular}  &
        \begin{tabular}{@{}c@{}} Speed \\ JSD $\times 10^{-2}$ \end{tabular}  &
        \begin{tabular}{@{}c@{}} Acc. \\ JSD $\times 10^{-2}$ \end{tabular}  &
        \begin{tabular}{@{}c@{}} $N_{LC}$ \\ JSD $\times 10^{-2}$ \end{tabular}  \\ \hline
\tdiffsimmse                            &                           1.21 ± 0.64 &       2.71 ± 1.41 &      1.69 ± 0.41 &    0.33 ± 0.27 &            2.22 ± 0.8 &                  0.45 ± 0.32 \\
\tdiffsimwmse                                 &                           0.63 ± 0.16 &       1.46 ± 0.38 &      1.69 ± 0.17 &    0.27 ± 0.06 &           1.57 ± 0.45 &                  0.63 ± 0.36 \\
\tdiffsimwmsecollision                     &                           \textbf{0.39 ± 0.17} &       \textbf{0.75 ± 0.39} &      1.83 ± 0.18 &     0.45 ± 0.2 &           4.79 ± 2.81 &                  0.88 ± 0.36 \\ \hline
\tmgailbcll (Gauss.)                  &                           2.65 ± 0.27 &       4.57 ± 0.67 &      2.14 ± 0.12 &    0.42 ± 0.16 &           2.54 ± 0.65 &                  1.62 ± 0.43 \\
\tmgailbcll (GMM)                       &                           1.88 ± 0.25 &       3.74 ± 0.16 &      2.16 ± 0.13 &     0.8 ± 0.22 &           4.35 ± 1.23 &                  0.64 ± 0.25 \\
\tmgaildiffsimwmse (Gauss.)             &                           0.65 ± 0.21 &        1.59 ± 0.5 &      1.16 ± 0.02 &    0.18 ± 0.05 &            3.1 ± 1.52 &                  \textbf{0.14 ± 0.09} \\
\tmgaildiffsimwmse (GMM)                  &                           0.44 ± 0.31 &       1.12 ± 0.59 &      \textbf{1.14 ± 0.09} &     \textbf{0.11 ± 0.0} &           \textbf{1.11 ± 0.25} &                   0.2 ± 0.14 \\
\tmgaildiffwmsecol (Gauss.) &                           1.28 ± 0.13 &       2.23 ± 0.47 &      1.59 ± 0.12 &    0.41 ± 0.11 &           2.24 ± 0.23 &                   0.8 ± 0.18 \\
\tmgaildiffwmsecol (GMM)      &                           0.54 ± 0.37 &        1.34 ± 0.5 &      1.46 ± 0.16 &    0.18 ± 0.04 &            1.41 ± 0.5 &                   0.4 ± 0.33 \\
\end{tabular}

    \label{tab:train_on_one_eval_on_all}
\end{table*}

\section{Analysis of Results}
\label{sec:discussion}
We analyze the experimental results and draw high-level conclusions for driver model training:

\textit{Closed-loop can be beneficial over open-loop training}:
Theoretical and experimental findings have been established in the past regarding the benefit of closed-loop multi-agent training over simpler, more open-loop approaches, e.g., the ``compounding error'' phenomenon of BC training \cite{ho2016generative,ross2010efficient,ross2011reduction}. 
Nonetheless, open-loop approaches are still often used due to their simplicity \cite{bansal2018chauffeurnet,igl2022symphony}.
Our experimental findings confirm the case made for closed-loop training. From Table \ref{tab:train_on_all_eval_on_all}, we see that the two closed-loop paradigms - differentiable simulation and MGAIL - significantly outperform open-loop BC methods. 
This is evident when comparing \ttdiffsim to \ttbcwMSE as well as \ttmgailbcll to the respective \texttt{BC-LL} method with the same head. 
In all cases, the closed-loop method is better or equal on (almost) every metric.
Additionally, single-agent closed-loop training with log replay of surrounding agents (Table \ref{tab:train_on_one_eval_on_all}) can be seen as ``less closed-loop'' than full multi-agent closed-loop training (Table \ref{tab:train_on_all_eval_on_all}). 
Also, here, the former outperforms the latter.
Furthermore, we can see that the closed loop methods lead to more realistic scenarios in terms of JSD metrics and and almost always to lower collision rates than an \tidm model.

\textit{Reinforcement loss can harm realism}:
Unrealistically high collision rates remain one of the biggest open challenges in learned driver models, and are a key indicator of where realism is still limited \cite{bronstein2022hierarchical, igl2022symphony}. As a natural remedy, various works \cite{suo2021trafficsim,bhattacharyya2019simulating} have added a ``common sense loss'', also called ``reinforcement loss''.
Our experiments show that such a reinforcement learning aspect can indeed significantly bring down the collision rate (e.g., compare collision rate metric between \ttdiffsimwmse and \ttdiffsimwmsecollision or between the respective MGAIL methods). 
However, this can come at the substantial cost of losing realism in other aspects of the model. 
For example, all three JSD metrics (speed, acceleration, $N_{LC}$) are worse (or equal) when comparing \ttdiffsimwmsecollision \textbf{vs.} \ttdiffsimwmse or considering \ttmgaildiffsimwmsecollision \textbf{vs.} \ttmgaildiffsimwmse. 
Here, the collision loss not only weakens the JSD metrics, but also has a negative effect on the off-road rate.

\textit{Deterministic supervised learning can be competitive with MGAIL}:
Recent work \cite{igl2022symphony,bhattacharyya2019simulating} increasingly utilized adversarial learning to train driver models. 
It has the considerable advantage that this allows for probabilistic policies, and the loss is learned instead of being defined manually.
However, adversarial methods come with training challenges and are usually less stable to train compared to supervised or variational inference methods \cite{bond2021deep}. 
In our experiments, we observe that methods employing \ttdiffsim outperform \ttmgailbcll on all metrics, except of the acceleration JSD. 
While both methods are trained closed-loop and improve upon their open-loop BC pretrainings, we observe a stronger improvement in the metrics of \ttdiffsimmse, \ttdiffsimwmse, and \ttdiffsimwmsecollision. 
However, we believe that this statement is dependent on the tuning of the MGAIL method as well as the probabilistic characteristics of the dataset, and should therefore be taken with caution. 
In particular, it might be the case that highway scenarios, including on-ramp and off-ramp situations, are not very multi-modal and thus a deterministic method is well suited.

\textit{Combinations of methods can be beneficial}: One key insight of our experiments is that combinations of methods can counteract individual weaknesses. 
We already observed that a collision loss can reduce collision rate, at the price of having worse distributional realism. 
However, adding adversarial learning helps mitigate this effect, as can be seen when comparing \ttdiffsimwmsecollision with \ttmgaildiffsimwmsecollision, where distributional realism is improved over all metrics. 
Furthermore, \ttmgaildiffsimwmse showed to be the best performing model in all metrics except of collision rate. 
This holds for combined as well as log-replay training. We think that the supervised learning signal helps to stabilize the adversarial training.

\section{Conclusion}
To summarize, we conducted a systematic analysis study of closed-loop imitation training principles for realistic traffic agent models for highway scenarios. 
We utilized the same GNN-based driving policy under different training paradigms,
and reported quantitative experimental results as well as high-level insights with qualitative results given in the supplementary material. 
We find that each method on its own comes with individual weaknesses, and combining different methods can counteract them. 
In particular, we find that closed-loop training has significant advantages over open-loop training, that a reinforcement signal can destroy realism and that combinations of different closed-loop learning principles can improve overall performance.

\bibliographystyle{unsrt}
\bibliography{references.bib}  

\begin{thebibliography}{10}

\bibitem{suo2021trafficsim}
Simon Suo, Sebastian Regalado, Sergio Casas, and Raquel Urtasun.
\newblock Trafficsim: Learning to simulate realistic multi-agent behaviors.
\newblock In {\em {IEEE} Conference on Computer Vision and Pattern Recognition, {CVPR} 2021, virtual, June 19-25, 2021}, pages 10400--10409. Computer Vision Foundation / {IEEE}, 2021.

\bibitem{bergamini2021simnet}
Luca Bergamini, Yawei Ye, Oliver Scheel, Long Chen, Chih Hu, Luca Del~Pero, Błażej Osiński, Hugo Grimmet, and Peter Ondruska.
\newblock Simnet: Learning reactive self-driving simulations from real-world observations.
\newblock In {\em 2021 IEEE International Conference on Robotics and Automation (ICRA)}, pages~--. IEEE, 2021.

\bibitem{driggs2017integrating}
Katherine Driggs-Campbell, Vijay Govindarajan, and Ruzena Bajcsy.
\newblock Integrating intuitive driver models in autonomous planning for interactive maneuvers.
\newblock {\em IEEE Transactions on Intelligent Transportation Systems}, 18(12):3461--3472, 2017.

\bibitem{feng2023trafficgen}
Lan Feng, Quanyi Li, Zhenghao Peng, Shuhan Tan, and Bolei Zhou.
\newblock Trafficgen: Learning to generate diverse and realistic traffic scenarios.
\newblock In {\em 2023 IEEE International Conference on Robotics and Automation (ICRA)}, pages 3567--3575. IEEE, 2023.

\bibitem{scheel2021urban}
Oliver Scheel, Luca Bergamini, Maciej Wolczyk, Blazej Osinski, and Peter Ondruska.
\newblock Urban driver: Learning to drive from real-world demonstrations using policy gradients.
\newblock In Aleksandra Faust, David Hsu, and Gerhard Neumann, editors, {\em Conference on Robot Learning, 8-11 November 2021, London, {UK}}, volume 164 of {\em Proceedings of Machine Learning Research}, pages 718--728. {PMLR}, 2021.

\bibitem{pmlr-v205-karkus23a}
Peter Karkus, Boris Ivanovic, Shie Mannor, and Marco Pavone.
\newblock Diffstack: A differentiable and modular control stack for autonomous vehicles.
\newblock In Karen Liu, Dana Kulic, and Jeff Ichnowski, editors, {\em Proceedings of The 6th Conference on Robot Learning}, volume 205 of {\em Proceedings of Machine Learning Research}, pages 2170--2180. PMLR, 14--18 Dec 2023.

\bibitem{Suo_2023_CVPR}
Simon Suo, Kelvin Wong, Justin Xu, James Tu, Alexander Cui, Sergio Casas, and Raquel Urtasun.
\newblock Mixsim: A hierarchical framework for mixed reality traffic simulation.
\newblock In {\em Proceedings of the IEEE/CVF Conference on Computer Vision and Pattern Recognition (CVPR)}, pages 9622--9631, June 2023.

\bibitem{kuefler2017imitating}
Alex Kuefler, Jeremy Morton, Tim Wheeler, and Mykel Kochenderfer.
\newblock Imitating driver behavior with generative adversarial networks.
\newblock In {\em 2017 IEEE intelligent vehicles symposium (IV)}, pages 204--211. IEEE, 2017.

\bibitem{tabatabaie2023interaction}
Mahan Tabatabaie, Suining He, and Kang~G Shin.
\newblock Interaction-aware and hierarchically-explainable heterogeneous graph-based imitation learning for autonomous driving simulation.
\newblock In {\em 2023 IEEE/RSJ International Conference on Intelligent Robots and Systems (IROS)}, pages 3576--3581. IEEE, 2023.

\bibitem{igl2022symphony}
Maximilian Igl, Daewoo Kim, Alex Kuefler, Paul Mougin, Punit Shah, Kyriacos Shiarlis, Dragomir Anguelov, Mark Palatucci, Brandyn White, and Shimon Whiteson.
\newblock Symphony: Learning realistic and diverse agents for autonomous driving simulation.
\newblock In {\em 2022 International Conference on Robotics and Automation (ICRA)}, pages 2445--2451. IEEE, 2022.

\bibitem{scibior2021itra}
Adam {\'{S}}cibior, Vasileios Lioutas, Daniele Reda, Peyman Bateni, and Frank Wood.
\newblock Imagining the road ahead: Multi-agent trajectory prediction via differentiable simulation.
\newblock In {\em 24th {IEEE} International Intelligent Transportation Systems Conference, {ITSC} 2021, Indianapolis, IN, USA, September 19-22, 2021}, pages 720--725. {IEEE}, 2021.

\bibitem{lioutas2022titrated}
Vasileios Lioutas, Adam Scibior, and Frank Wood.
\newblock Titrated: Learned human driving behavior without infractions via amortized inference.
\newblock {\em Transactions on Machine Learning Research}, 2022.

\bibitem{zhang2023learning}
Chris Zhang, James Tu, Lunjun Zhang, Kelvin Wong, Simon Suo, and Raquel Urtasun.
\newblock Learning realistic traffic agents in closed-loop.
\newblock In {\em 7th Annual Conference on Robot Learning}, 2023.

\bibitem{xu2023bits}
Danfei Xu, Yuxiao Chen, Boris Ivanovic, and Marco Pavone.
\newblock Bits: Bi-level imitation for traffic simulation.
\newblock In {\em 2023 IEEE International Conference on Robotics and Automation (ICRA)}, pages 2929--2936. IEEE, 2023.

\bibitem{lu2023imitation}
Yiren Lu, Justin Fu, George Tucker, Xinlei Pan, Eli Bronstein, Rebecca Roelofs, Benjamin Sapp, Brandyn White, Aleksandra Faust, Shimon Whiteson, et~al.
\newblock Imitation is not enough: Robustifying imitation with reinforcement learning for challenging driving scenarios.
\newblock In {\em 2023 IEEE/RSJ International Conference on Intelligent Robots and Systems (IROS)}, pages 7553--7560. IEEE, 2023.

\bibitem{bronstein2022hierarchical}
Eli Bronstein, Mark Palatucci, Dominik Notz, Brandyn White, Alex Kuefler, Yiren Lu, Supratik Paul, Payam Nikdel, Paul Mougin, Hongge Chen, et~al.
\newblock Hierarchical model-based imitation learning for planning in autonomous driving.
\newblock In {\em 2022 IEEE/RSJ International Conference on Intelligent Robots and Systems (IROS)}, pages 8652--8659. IEEE, 2022.

\bibitem{moers2022exid}
Tobias Moers, Lennart Vater, Robert Krajewski, Julian Bock, Adrian Zlocki, and Lutz Eckstein.
\newblock The exid dataset: A real-world trajectory dataset of highly interactive highway scenarios in germany.
\newblock In {\em 2022 IEEE Intelligent Vehicles Symposium (IV)}, pages 958--964, 2022.

\bibitem{treiber2000congested}
Martin Treiber, Ansgar Hennecke, and Dirk Helbing.
\newblock Congested traffic states in empirical observations and microscopic simulations.
\newblock {\em Physical review E}, 62(2):1805, 2000.

\bibitem{kesting2010enhanced}
Arne Kesting, Martin Treiber, and Dirk Helbing.
\newblock Enhanced intelligent driver model to access the impact of driving strategies on traffic capacity.
\newblock {\em Philosophical Transactions of the Royal Society A: Mathematical, Physical and Engineering Sciences}, 368(1928):4585--4605, 2010.

\bibitem{zhou2016impact}
Mofan Zhou, Xiaobo Qu, and Sheng Jin.
\newblock On the impact of cooperative autonomous vehicles in improving freeway merging: a modified intelligent driver model-based approach.
\newblock {\em IEEE Transactions on Intelligent Transportation Systems}, 18(6):1422--1428, 2016.

\bibitem{sharath2020enhanced}
MN~Sharath and Nagendra~R Velaga.
\newblock Enhanced intelligent driver model for two-dimensional motion planning in mixed traffic.
\newblock {\em Transportation Research Part C: Emerging Technologies}, 120:102780, 2020.

\bibitem{kanagaraj2013evaluation}
Venkatesan Kanagaraj, Gowri Asaithambi, CH~Naveen Kumar, Karthik~K Srinivasan, and R~Sivanandan.
\newblock Evaluation of different vehicle following models under mixed traffic conditions.
\newblock {\em Procedia-Social and Behavioral Sciences}, 104:390--401, 2013.

\bibitem{pourabdollah2017calibration}
Mitra Pourabdollah, Eric Bj{\"a}rkvik, Florian F{\"u}rer, Bj{\"o}rn Lindenberg, and Klaas Burgdorf.
\newblock Calibration and evaluation of car following models using real-world driving data.
\newblock In {\em 2017 IEEE 20th International conference on intelligent transportation systems (ITSC)}, pages 1--6. IEEE, 2017.

\bibitem{zhang2021comprehensive}
Duo Zhang, Xiaoyun Chen, Junhua Wang, Yinhai Wang, and Jian Sun.
\newblock A comprehensive comparison study of four classical car-following models based on the large-scale naturalistic driving experiment.
\newblock {\em Simulation Modelling Practice and Theory}, 113:102383, 2021.

\bibitem{wiedemann1974simulation}
Rainer Wiedemann.
\newblock Simulation des strassenverkehrsflusses.
\newblock 1974.

\bibitem{chakroborty1999evaluation}
Partha Chakroborty and Shinya Kikuchi.
\newblock Evaluation of the general motors based car-following models and a proposed fuzzy inference model.
\newblock {\em Transportation Research Part C: Emerging Technologies}, 7(4):209--235, 1999.

\bibitem{krauss1997metastable}
Stefan Krau{\ss}, Peter Wagner, and Christian Gawron.
\newblock Metastable states in a microscopic model of traffic flow.
\newblock {\em Physical Review E}, 55(5):5597, 1997.

\bibitem{gipps1981behavioural}
Peter~G Gipps.
\newblock A behavioural car-following model for computer simulation.
\newblock {\em Transportation Research Part B: Methodological}, 15(2):105--111, 1981.

\bibitem{kesting2007general}
Arne Kesting, Martin Treiber, and Dirk Helbing.
\newblock General lane-changing model mobil for car-following models.
\newblock {\em Transportation Research Record}, 1999(1):86--94, 2007.

\bibitem{teng2023motion}
Siyu Teng, Xuemin Hu, Peng Deng, Bai Li, Yuchen Li, Yunfeng Ai, Dongsheng Yang, Lingxi Li, Zhe Xuanyuan, Fenghua Zhu, et~al.
\newblock Motion planning for autonomous driving: The state of the art and future perspectives.
\newblock {\em IEEE Transactions on Intelligent Vehicles}, 2023.

\bibitem{chen2024data}
Di~Chen, Meixin Zhu, Hao Yang, Xuesong Wang, and Yinhai Wang.
\newblock Data-driven traffic simulation: A comprehensive review.
\newblock {\em IEEE Transactions on Intelligent Vehicles}, 2024.

\bibitem{shiroshita2020behaviorally}
Shinya Shiroshita, Shirou Maruyama, Daisuke Nishiyama, Mario~Ynocente Castro, Karim Hamzaoui, Guy Rosman, Jonathan DeCastro, Kuan-Hui Lee, and Adrien Gaidon.
\newblock Behaviorally diverse traffic simulation via reinforcement learning.
\newblock In {\em 2020 IEEE/RSJ International Conference on Intelligent Robots and Systems (IROS)}, pages 2103--2110. IEEE, 2020.

\bibitem{Chen_2021_ICCV}
Dian Chen, Vladlen Koltun, and Philipp Kr\"ahenb\"uhl.
\newblock Learning to drive from a world on rails.
\newblock In {\em Proceedings of the IEEE/CVF International Conference on Computer Vision (ICCV)}, pages 15590--15599, October 2021.

\bibitem{Toromanoff_2020_CVPR}
Marin Toromanoff, Emilie Wirbel, and Fabien Moutarde.
\newblock End-to-end model-free reinforcement learning for urban driving using implicit affordances.
\newblock In {\em Proceedings of the IEEE/CVF Conference on Computer Vision and Pattern Recognition (CVPR)}, June 2020.

\bibitem{fu2017learning}
Justin Fu, Katie Luo, and Sergey Levine.
\newblock Learning robust rewards with adversarial inverse reinforcement learning.
\newblock {\em arXiv preprint arXiv:1710.11248}, 2017.

\bibitem{zheng2021objective}
Guanjie Zheng, Hanyang Liu, Kai Xu, and Zhenhui Li.
\newblock Objective-aware traffic simulation via inverse reinforcement learning.
\newblock {\em arXiv preprint arXiv:2105.09560}, 2021.

\bibitem{wu2020efficient}
Zheng Wu, Liting Sun, Wei Zhan, Chenyu Yang, and Masayoshi Tomizuka.
\newblock Efficient sampling-based maximum entropy inverse reinforcement learning with application to autonomous driving.
\newblock {\em IEEE Robotics and Automation Letters}, 5(4):5355--5362, 2020.

\bibitem{bansal2018chauffeurnet}
Mayank Bansal, Alex Krizhevsky, and Abhijit Ogale.
\newblock Chauffeurnet: Learning to drive by imitating the best and synthesizing the worst.
\newblock {\em arXiv preprint arXiv:1812.03079}, 2018.

\bibitem{ross2011reduction}
St{\'e}phane Ross, Geoffrey Gordon, and Drew Bagnell.
\newblock A reduction of imitation learning and structured prediction to no-regret online learning.
\newblock In {\em Proceedings of the fourteenth international conference on artificial intelligence and statistics}, pages 627--635. JMLR Workshop and Conference Proceedings, 2011.

\bibitem{gao2020vectornet}
Jiyang Gao, Chen Sun, Hang Zhao, Yi~Shen, Dragomir Anguelov, Congcong Li, and Cordelia Schmid.
\newblock Vectornet: Encoding hd maps and agent dynamics from vectorized representation.
\newblock In {\em Proceedings of the IEEE/CVF Conference on Computer Vision and Pattern Recognition}, pages 11525--11533, 2020.

\bibitem{song2016mgail}
Jiaming Song, Hongyu Ren, Dorsa Sadigh, and Stefano Ermon.
\newblock Multi-agent generative adversarial imitation learning.
\newblock In S.~Bengio, H.~Wallach, H.~Larochelle, K.~Grauman, N.~Cesa-Bianchi, and R.~Garnett, editors, {\em Advances in Neural Information Processing Systems}, volume~31. Curran Associates, Inc., 2018.

\bibitem{pmlr-v80-mescheder18a}
Lars Mescheder, Andreas Geiger, and Sebastian Nowozin.
\newblock Which training methods for {GAN}s do actually converge?
\newblock In Jennifer Dy and Andreas Krause, editors, {\em Proceedings of the 35th International Conference on Machine Learning}, volume~80 of {\em Proceedings of Machine Learning Research}, pages 3481--3490. PMLR, 10--15 Jul 2018.

\bibitem{dadashi2020primal}
Robert Dadashi, L{\'e}onard Hussenot, Matthieu Geist, and Olivier Pietquin.
\newblock Primal wasserstein imitation learning.
\newblock {\em arXiv preprint arXiv:2006.04678}, 2020.

\bibitem{yan2023learning}
Xintao Yan, Zhengxia Zou, Shuo Feng, Haojie Zhu, Haowei Sun, and Henry~X Liu.
\newblock Learning naturalistic driving environment with statistical realism.
\newblock {\em Nature communications}, 14(1):2037, 2023.

\bibitem{casas20spagnn}
S.~{Casas}, C.~{Gulino}, R.~{Liao}, and R.~{Urtasun}.
\newblock Spagnn: Spatially-aware graph neural networks for relational behavior forecasting from sensor data.
\newblock In {\em 2020 IEEE International Conference on Robotics and Automation (ICRA)}, pages 9491--9497, 2020.

\bibitem{ho2016generative}
Jonathan Ho and Stefano Ermon.
\newblock Generative adversarial imitation learning.
\newblock {\em Advances in neural information processing systems}, 29, 2016.

\bibitem{ross2010efficient}
St{\'e}phane Ross and Drew Bagnell.
\newblock Efficient reductions for imitation learning.
\newblock In {\em Proceedings of the thirteenth international conference on artificial intelligence and statistics}, pages 661--668. JMLR Workshop and Conference Proceedings, 2010.

\bibitem{bhattacharyya2019simulating}
Raunak~P Bhattacharyya, Derek~J Phillips, Changliu Liu, Jayesh~K Gupta, Katherine Driggs-Campbell, and Mykel~J Kochenderfer.
\newblock Simulating emergent properties of human driving behavior using multi-agent reward augmented imitation learning.
\newblock In {\em 2019 International Conference on Robotics and Automation (ICRA)}, pages 789--795. IEEE, 2019.

\bibitem{bond2021deep}
Sam Bond-Taylor, Adam Leach, Yang Long, and Chris~G Willcocks.
\newblock Deep generative modelling: A comparative review of vaes, gans, normalizing flows, energy-based and autoregressive models.
\newblock {\em IEEE transactions on pattern analysis and machine intelligence}, 44(11):7327--7347, 2021.

\bibitem{poggenhans2018lanelet2}
Fabian Poggenhans, Jan-Hendrik Pauls, Johannes Janosovits, Stefan Orf, Maximilian Naumann, Florian Kuhnt, and Matthias Mayr.
\newblock Lanelet2: A high-definition map framework for the future of automated driving.
\newblock In {\em Proc.\ IEEE Intell.\ Trans.\ Syst.\ Conf.}, Hawaii, USA, November 2018.

\bibitem{shi2020masked}
Yunsheng Shi, Zhengjie Huang, Shikun Feng, Hui Zhong, Wenjin Wang, and Yu~Sun.
\newblock Masked label prediction: Unified message passing model for semi-supervised classification.
\newblock {\em arXiv preprint arXiv:2009.03509}, 2020.

\end{thebibliography}




\section*{APPENDIX}

\subsection{Architecture overview}
\label{app:arch_overview}
\begin{figure*}[h]
    \centering
    \includegraphics[width=0.95\linewidth]{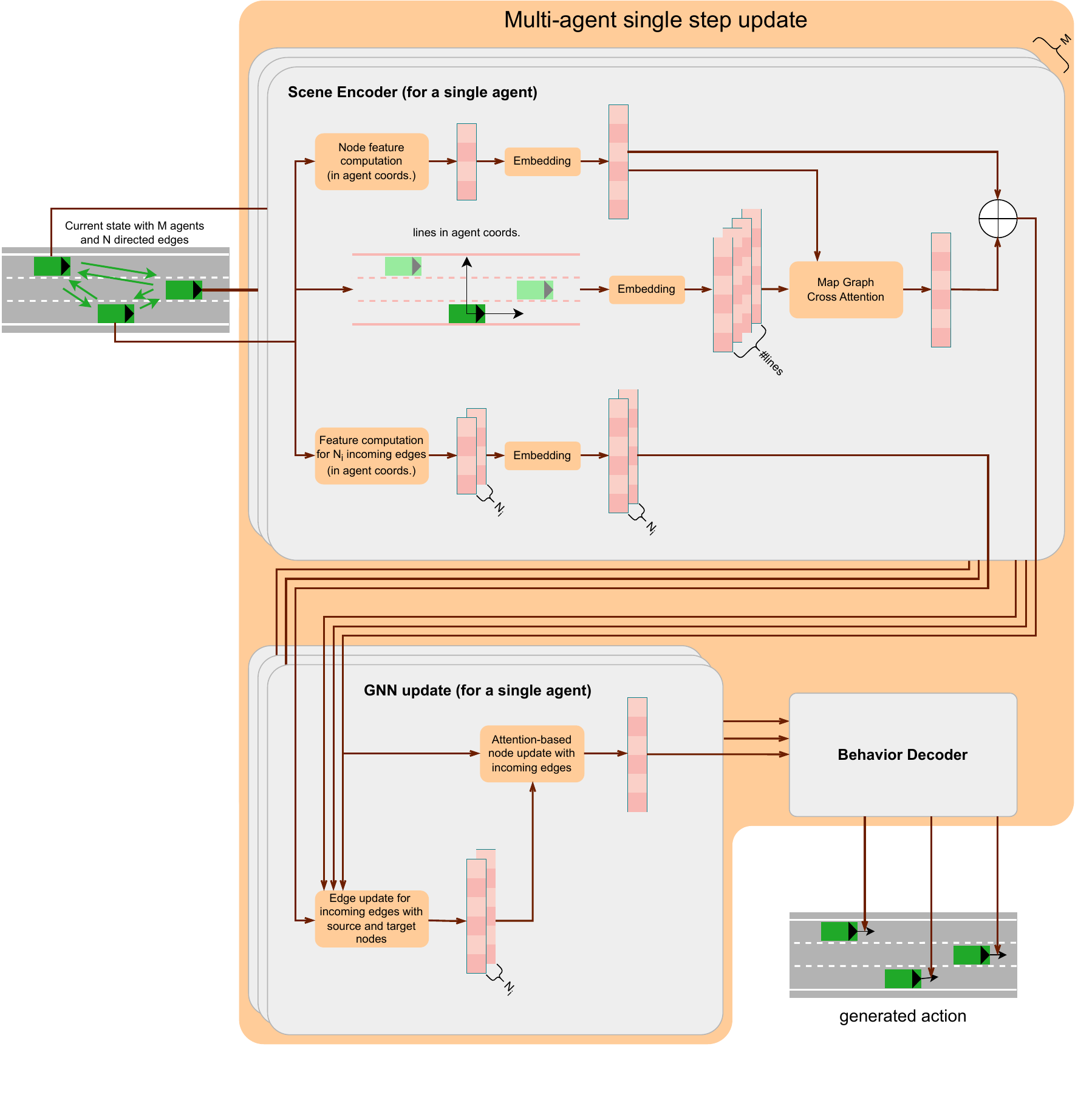}
    \caption{Architecture Overview}
    \label{fig:gnn_architecture}
\end{figure*}

In Section \ref{sec:method}, we gave a formal introduction to our policy parameterization. In the following, we add an intuitive description of our architecture along with a visualization that can be seen in Figure \ref{fig:gnn_architecture}.

To compute the action for each agent in a given scene, we employ a \emph{Scene Encoder}, followed by a \emph{GNN}, followed by a \emph{Behavior Decoder}. As can be seen in Figure \ref{fig:gnn_architecture}, the input to the network is a scene at a certain point in time. It contains a map, which consists of lane graph information and lane boundaries and it contains features for each agent like the position, bounding box, orientation, velocity, acceleration. Based on these features we determine, for each pair of agents, whether one is relevant for the action of the other. If that is the case, we introduce a directed edge between them in our GNN.

Based on this input, the \emph{Scene Encoder} computes node and edge features (like the relative position or velocity of two agents) and embeds them in latent space. Moreover, for each agent, the lines in the relevant part of the map are extracted, transformed into ego coordinates and embedded into latent space. After that, a cross attention module (which also takes the node embedding into account) computes a map embedding from the line embeddings (see Figure \ref{fig:map embedder} for details). 
The map embedding is added to the node embedding as the last step of the Scene Encoder. Hence, the output of the Scene Encoder is an embedding for each agent and an embedding for each directed edge between two agents.

These embeddings enter the \emph{GNN} that computes one update of the edge embedding by taking into account the node embeddings of the source and the target node. Afterwards, each node is updated by taking into account all embeddings of incoming edges. This yields a final embedding for each agent that is passed to the \emph{Behavior Decoder}. This is an MLP that generates the action in the desired output format (e.g., means and covariances of the Gaussians or next waypoints).

\subsubsection{Polyline representation.}
\begin{figure*}[h]
    \centering
    \includegraphics[width=1\linewidth]{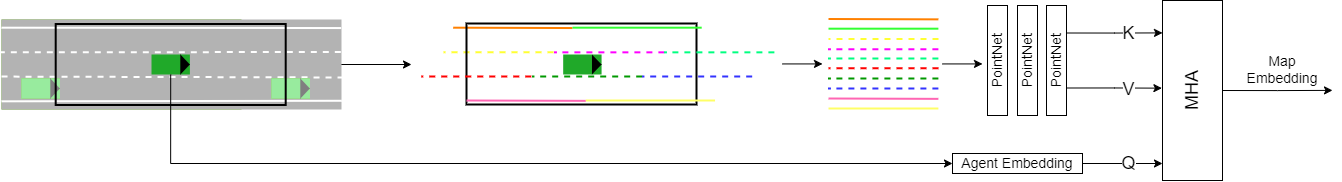}
    \caption{Road topology extraction and embedding. }
    \label{fig:map embedder}
\end{figure*}

To embed the map information, we use a similar method as in \cite{gao2020vectornet}. 
Road lines are extracted from Lanelet2 \cite{poggenhans2018lanelet2} map representation and split into segments of maximum length of 20 meters.
Each line segment is represented as a polyline of 10 points.
Segments are selected that fall into a crop around each agent in each respective agent's frame of reference with at least one point.
The crop size is 10 meters to left and to right, 120 meters in front and 45 meters in the rear of the vehicle.
Selected polylines are combined with a sinusoidal embedding and the line type.
We use two line types - solid and dashed.
Each polyline is passed through three PointNet \cite{gao2020vectornet} layers.
Embedded polylines are used as key and value arguments in Multi-Head Attention (MHA) message passing module (as implemented in \cite{shi2020masked}) where query is the relevant agent embedding. 
Aggregated attended polyline embeddings are then combined with the agent embedding to obtain the agent representation.
Map topology embedding is depicted in Figure \ref{fig:map embedder}.

\subsection{Method Details + Hyperparameters} \label{subsec: method details}
As default optimizer, we use Adam with learning rate $l_{\textrm{rate}}$ and a StepLR learning rate scheduler with factor $\gamma$ and step-size $n_\textrm{step}$. 
These three parameters were tuned independently for combined and log-replay trainings. 
Here, we only report them for the combined trainings - all other hyperparameters are the same for both settings. 
Furthermore, when we use a weighted MSE we mean for $(x,y)\in\mathbb{R}^{2}$
\begin{align*}
    d_{\textrm{wMSE}}((\hat{x},\hat{y}),(x,y))=\alpha_{x}d_{\textrm{MSE}}(\hat{x},x)+\alpha_{y}d_{\textrm{MSE}}(\hat{y},y).
\end{align*}

\textbf{\tbcgaussll/\tbcgmmll.} 
For MGAIL based methods, we pretrain and compare with BC methods that share the same head (Gaussian or GMM) and that are trained with maximum likelihood. Thus, for a given policy parameterization $\pi_{\theta}(\mathbf{a}|\mathbf{s})$ we minimize
\begin{align*}
    \mathcal{L}_{\mathrm{BC}}  (\theta)=\mathbb{E}_{\mathbf{s},\mathbf{a}\sim\mathcal{D}}[-\mathrm{log}\ \pi_{\theta}(\mathbf{a}|\mathbf{s})].
\end{align*}
For \ttbcgaussll, a hyperparameter search determined a learning rate of $l_{\textrm{rate}}=0.001$, with scheduler parameters $\gamma=0.95$ and $n_\textrm{step}=2$ and for \ttbcgmmll, it results in $l_{\textrm{rate}}=0.001$, $\gamma=0.99$ and $n_\textrm{step}=2$.

\textbf{\tbcwMSE.} Our BC method, that we use for comparison and pretraining for the deterministic differentiable simulation methods, is trained via minimizing
\begin{align*}
     \mathcal{L}_{\mathrm{BC}}  (\theta)=\mathbb{E}_{\mathbf{s},\mathbf{a}\sim\mathcal{D}}[d_{\textrm{wMSE-Pos}}(\hat{\mathbf{a}},\mathbf{a})+\beta d_{\textrm{MSE-Orientation}}(\hat{\mathbf{a}},\mathbf{a}))],
\end{align*}
where $\hat{\mathbf{a}}=\pi_{\theta}(\mathbf{s})$, and
\begin{align*}
    d_{\textrm{wMSE-Pos}}(\hat{\mathbf{a}},\mathbf{a}):&=\frac{1}{M}\sum_{i=1}^{M}d_{\textrm{wMSE}}((\Delta\hat{x}_{i},\Delta\hat{y}_{i}),(\Delta x_{i},\Delta y_{i}))
\end{align*}
and 
\begin{align*}
    d_{\textrm{MSE-Orientation}}(\hat{\mathbf{a}},\mathbf{a}):=\frac{1}{M}\sum_{i=1}^{M}d_{\textrm{MSE}}((\hat{\delta}_{x,i},\hat{\delta}_{y,i}),(\delta_{x,i},\delta_{y,i})),
\end{align*}
where $(\delta_{x,i},\delta_{y,i})$ is the ground-truth heading of the next time step for agent $i$ and $(\hat{\delta}_{x,i},\hat{\delta}_{y,i})$ is the heading that would result from the chosen action. 
The loss weights used for \ttbcwMSE are $\alpha_{x}=0.10472$, $\alpha_{y}=65.177$ and $\beta=6209.8$ and are found using inverse-variance weighting. The learning rate hyperparameters are $l_{\textrm{rate}}=0.0005$, $n_\textrm{step}=1$ and $\gamma=0.99$. 
Further ablation studies of BC methods can be found in Section \ref{subsec:bc_ablation}.

\textbf{\tdiffsimwmse.} In differentiable simulation, we roll out the policy in closed-loop and minimize
\begin{align*}
    \mathcal{L}_{\mathrm{DS}}(\theta)=\mathbb{E}_{\mathbf{s}_{0}\sim\mathcal{D}}\bigg[\sum_{t=1}^{T}d(\mathbf{s}_{t},\tilde{\mathbf{s}}_{t})\bigg]
\end{align*}
with 
\begin{align*}
    d(\mathbf{s}_{t},\tilde{\mathbf{s}}_{t})=\frac{1}{M}\sum_{i=1}^{M}d_{\textrm{wMSE}}(g_{\mathrm{local}}(\hat{x}_{i,t},\hat{y}_{i,t}),g_{\mathrm{local}}(x_{i,t},y_{i,t})),
\end{align*}
where $(\hat{x}_{i,t},\hat{y}_{i,t})$, $(x_{i,t},y_{i,t})$ are generated and ground-truth global positions of agent $i$ in time step $t$ and $g_{\mathrm{local}}:\mathbb{R}^{2}\to\mathbb{R}^{2}$ transforms the global positions into the local coordinate system with origin $(x_{i,t-1},y_{i,t-1})$ and heading vector $(\delta_{x,i,t-1},\delta_{y,i,t-1})$ as x-axis.
The loss weights used for \ttdiffsimwmse are $\alpha_{x}=0.10472$, $\alpha_{y}=65.177$ and are found using inverse-variance weighting.

\textbf{Collision loss.} As collision loss, we use the circle-based differentiable relaxation of a collision, presented in \cite{suo2021trafficsim}, configured with a five circle representation for each vehicle. We denote this loss with $\mathcal{L}_\textrm{Collision}(\theta)$.

\textbf{\tdiffsimwmsecollision.} Here, we use as loss function
\begin{align*}
    \mathcal{L}_{DS+Col}(\theta)=\mathcal{L}_\mathrm{DS}(\theta)+\beta\mathcal{L}_{\mathrm{Collision}}(\theta).
\end{align*}
As loss hyperparameters, we use $\alpha_{x}=0.1$ and $\alpha_y=2.8$ inside the \ttdiffsimwmse loss and set $\beta=4.0$. 
For the learning rate hyperparameters, we use $l_{\textrm{rate}}=2.5e-05$, $\gamma=0.99$ and $n_\textrm{step}=2$.

\textbf{\tmgailbcll.} As described in Section \ref{sec:training}, in MGAIL a discriminator $D_{\psi}$ is trained alongside the generator $\pi_{\theta}$. The discriminator is trained via minimizing
\begin{align*}
    \mathcal{L}(\psi)=\mathbb{E}_{\mathbf{s}\sim{D}}[-\mathrm{log}\, D_{\psi}(\mathbf{s})]+\mathbb{E}_{\mathbf{s}\sim \pi_{\theta}} [-\mathrm{log}\, (1-D_{\psi}(\mathbf{s}))].
\end{align*}
To stabilize the generator training, it is commonly combined with the BC loss (see \cite{bronstein2022hierarchical}), which we also do here via
\begin{align*}
    \mathcal{L}(\theta)=\alpha\mathbb{E}_{\mathbf{s}\sim \pi_{\theta}} [\mathrm{log}\, (1-D_{\psi}(\mathbf{s}))]+\beta\mathbb{E}_{\mathbf{s},\mathbf{a}\sim\mathcal{D}}[-\mathrm{log}\ \pi_{\theta}(\mathbf{a}|\mathbf{s})].
\end{align*}
We note here, that the MGAIL loss is computed in closed-loop, meaning that the gradient can propagate back through time, whereas the BC loss is computed open-loop. Discriminator and generator have different loss functions and therefore also need a different learning rate. Our loss hyperparameters are $\alpha=50.0$ and $\beta=1.0$. For the Gaussian head, we set the learning rate for the discriminator to $l_{\textrm{rate}}=2e-05$ and for the generator to $l_{\textrm{rate}}=2e-05$. In both cases, we used $\gamma=0.99$ and $n_\textrm{step}=2$. For the Gaussian mixture head, we set the discriminator learning rate to $l_{\textrm{rate}}=1e-04$ and the generator learning rate to $l_{\textrm{rate}}=5e-05$. In both cases, we also used $\gamma=0.5$ and $n_\textrm{step}=20$.

\textbf{\tmgaildiffsimwmse.} For the combination of MGAIL with \ttdiffsimwmse, we change the generator loss to
\begin{align*}
    \mathcal{L}(\theta)=\alpha\mathbb{E}_{\mathbf{s}\sim \pi_{\theta}} [\mathrm{log}\, (1-D_{\psi}(\mathbf{s}))]+\beta\mathcal{L}_{\mathrm{DS}}(\theta).
\end{align*}
We use the standard loss weights of \ttdiffsimwmse and set $\alpha=\beta=1.0$. Furthermore, for the Gaussian head, we use as learning rate for the discriminator $l_{\textrm{rate}}=1e-04$ and for the generator $l_{\textrm{rate}}=5e-05$. In both cases, we used $\gamma=0.5$ and $n_\textrm{step}=40$. For the Gaussian mixture head, we set the discriminator learning rate to $l_{\textrm{rate}}=2e-04$ and the generator learning rate to $l_{\textrm{rate}}=1e-04$. In both cases, we also used $\gamma=0.5$ and $n_\textrm{step}=20$.

\textbf{\tmgaildiffsimwmsecollision.} For the additional combination with the collision loss and \ttmgaildiffsimwmse, we use as generator loss
\begin{align*}
    \mathcal{L}(\theta)=\alpha_{1}\mathbb{E}_{\mathbf{s}\sim \pi_{\theta}} [\mathrm{log}\, (1-D_{\psi}(\mathbf{s}))]+\beta\mathcal{L}_{\mathrm{DS+Col}}(\theta).
\end{align*}
For $\mathcal{L}_{\mathrm{DS+Col}}$ loss weights, we use the same as in \ttdiffsimwmsecollision and set $\alpha=5.0$ and $\beta=1.0$. Furthermore, for the Gaussian head we use as learning rate for the discriminator $l_{\textrm{rate}}=1e-03$ and for the generator $l_{\textrm{rate}}=5e-05$. 
In both cases we used $\gamma=0.5$ and $n_\textrm{step}=40$. For the Gaussian mixture head the setting is the same.

\subsection{Further Evaluations}

\subsubsection{Log-replay evaluation} \label{subsec: log-replay eval}
In Table \ref{tab:train_on_all_eval_on_all} and Table \ref{tab:train_on_one_eval_on_all}, we report results on different agent control methods when evaluated on controlling all agents in the scene. 
Here, we report evaluation on an inverse task of controlling only one agent in the scene, i.e. a learned agent policy executed alongside log-replay agents. 
The single agent that is controlled in the evaluation is deterministically selected as the agent that is present with the most time steps in the scene.
We train the policy in two settings - controlling all agents  and controlling a single agent along log-replayed agents during training.
The evaluation results are given in Table \ref{tab:train_on_all_eval_on_one} for training method with all agents and Table \ref{tab:train_on_one_eval_on_one} for a single agent.
While in all agent control evaluation the models trained alongside log-replay agents performed worse, we do not see such difference in performance here.
If controlling only a single agent alongside replayed agents in the simulation rollout both control methods in training perform approximately the same.
However, the introduction of collision loss no longer boosts the collision rate results.

\begin{table*}
    \centering
    \caption{Loss ablation study for multi-step closed-loop trainings when trained on controlling \textbf{all agents} and evaluated when controlling \textbf{one agent}. }
    \scriptsize
    \begin{tabular}{l|cccccc} 
        \hline
        Method  &
        \begin{tabular}{@{}c@{}} Col. \\ (\%) \end{tabular}  &
        \begin{tabular}{@{}c@{}} Off. \\ (\%) \end{tabular}  &
        \begin{tabular}{@{}c@{}} ADE \\ (m) \end{tabular}  &
        \begin{tabular}{@{}c@{}} Speed \\ JSD $\times 10^{-2}$ \end{tabular}  &
        \begin{tabular}{@{}c@{}} Acc. \\ JSD $\times 10^{-2}$ \end{tabular}  &
        \begin{tabular}{@{}c@{}} $N_{LC}$ \\ JSD $\times 10^{-2}$ \end{tabular}  \\ \hline
\tbcgaussll                        &                           0.13 ± 0.01 &        0.4 ± 0.06 &      1.51 ± 0.13 &    7.02 ± 2.21 &           6.69 ± 3.86 &                 11.69 ± 0.54 \\
\tbcgmmll                             &                           0.13 ± 0.04 &       0.47 ± 0.12 &      2.44 ± 0.37 &      \textbf{5.3 ± 3.9} &          22.67 ± 6.05 &                  10.5 ± 0.55 \\
\tbcwMSE                                    &                           0.16 ± 0.03 &       0.25 ± 0.04 &      2.57 ± 0.24 &    7.16 ± 2.72 &           11.32 ± 1.7 &                 10.74 ± 0.51 \\
\hline
\tdiffsimmse                           &                           0.05 ± 0.03 &        0.22 ± 0.3 &      1.55 ± 0.73 &    6.98 ± 0.73 &           2.78 ± 2.22 &                  4.77 ± 6.28 \\
\tdiffsimwmse                               &                           0.04 ± 0.01 &       \textbf{0.01 ± 0.01} &       1.28 ± 0.2 &    7.54 ± 0.29 &            2.89 ± 1.9 &                  \textbf{0.43 ± 0.25} \\
\tdiffsimwmsecollision                   &                           0.04 ± 0.02 &       0.15 ± 0.18 &       2.51 ± 1.4 &   11.84 ± 6.84 &           2.08 ± 0.33 &                  2.51 ± 1.84 \\ \hline
\tmgailbcll (Gauss.)                  &                           0.09 ± 0.03 &       0.09 ± 0.05 &      1.61 ± 0.14 &    6.14 ± 0.32 &           2.76 ± 0.95 &                  4.69 ± 2.99 \\
\tmgailbcll (GMM)                       &                           0.34 ± 0.33 &       0.09 ± 0.05 &       6.5 ± 7.91 &   13.4 ± 15.61 &          15.19 ± 8.83 &                  2.45 ± 1.57 \\
\tmgaildiffsimwmse (Gauss.)             &                           \textbf{0.02 ± 0.01} &       0.02 ± 0.01 &      \textbf{0.89 ± 0.04} &    6.95 ± 0.48 &           \textbf{1.26 ± 0.25} &                  0.99 ± 0.34 \\
\tmgaildiffsimwmse (GMM)                  &                           \textbf{0.02 ± 0.01} &       0.02 ± 0.01 &      0.93 ± 0.14 &    6.67 ± 0.46 &           1.73 ± 0.32 &                  1.11 ± 0.73 \\
\tmgaildiffwmsecol (Gauss.) &                           0.05 ± 0.02 &       0.09 ± 0.05 &      1.35 ± 0.12 &     6.9 ± 0.52 &           2.15 ± 0.55 &                   3.61 ± 1.5 \\
\tmgaildiffwmsecol (GMM)      &                            0.03 ± 0.0 &       0.07 ± 0.03 &      1.23 ± 0.13 &    7.26 ± 0.73 &           2.22 ± 0.74 &                  2.46 ± 1.42 \\
\end{tabular}

    \label{tab:train_on_all_eval_on_one}
\end{table*}

\begin{table*}
    \centering
    \caption{Loss ablation study for multi-step closed-loop trainings when trained on controlling \textbf{one agent} and evaluated when controlling \textbf{one agent}. }
    \scriptsize
    \begin{tabular}{l|cccccc} 
        \hline
        Method  &
        \begin{tabular}{@{}c@{}} Col. \\ (\%) \end{tabular}  &
        \begin{tabular}{@{}c@{}} Off. \\ (\%) \end{tabular}  &
        \begin{tabular}{@{}c@{}} ADE \\ (m) \end{tabular}  &
        \begin{tabular}{@{}c@{}} Speed \\ JSD $\times 10^{-2}$ \end{tabular}  &
        \begin{tabular}{@{}c@{}} Acc. \\ JSD $\times 10^{-2}$ \end{tabular}  &
        \begin{tabular}{@{}c@{}} $N_{LC}$ \\ JSD $\times 10^{-2}$ \end{tabular}  \\ \hline
\tdiffsimmse                           &                           0.04 ± 0.03 &       0.18 ± 0.12 &      1.57 ± 0.31 &    7.57 ± 1.23 &           2.69 ± 1.84 &                  2.32 ± 1.76 \\
\tdiffsimwmse                            &                           0.04 ± 0.01 &       0.05 ± 0.02 &       1.6 ± 0.19 &    7.77 ± 0.93 &           2.32 ± 0.68 &                   1.42 ± 0.5 \\
\tdiffsimwmsecollision                    &                           0.04 ± 0.01 &       0.04 ± 0.02 &      1.82 ± 0.16 &    7.46 ± 0.61 &           5.18 ± 2.65 &                   \textbf{0.6 ± 0.36} \\ \hline
\tmgailbcll (Gauss.)                  &                           0.12 ± 0.02 &       0.14 ± 0.05 &      1.72 ± 0.18 &    5.66 ± 1.04 &           3.91 ± 1.35 &                  7.67 ± 1.28 \\
\tmgailbcll (GMM)                       &                           0.07 ± 0.02 &       0.11 ± 0.04 &      1.57 ± 0.18 &    \textbf{4.84 ± 1.62} &            4.9 ± 2.89 &                  4.46 ± 1.28 \\
\tmgaildiffsimwmse (Gauss.)             &                           0.04 ± 0.01 &       0.04 ± 0.01 &      1.03 ± 0.04 &    7.57 ± 0.34 &           2.15 ± 0.77 &                  2.46 ± 0.83 \\
\tmgaildiffsimwmse (GMM)                  &                           \textbf{0.02 ± 0.01} &       \textbf{0.03 ± 0.01} &       \textbf{1.0 ± 0.04} &    6.73 ± 0.33 &           \textbf{1.44 ± 0.35} &                  2.09 ± 0.89 \\
\tmgaildiffwmsecol (Gauss.) &                           0.06 ± 0.01 &       0.11 ± 0.03 &      1.32 ± 0.11 &    7.36 ± 0.32 &           2.49 ± 0.32 &                  5.46 ± 0.49 \\
\tmgaildiffwmsecol (GMM) 
&                           0.03 ± 0.02 &       0.05 ± 0.02 &      1.24 ± 0.16 &    7.01 ± 0.41 &            1.9 ± 0.46 &                  2.37 ± 1.48 \\
\end{tabular}

    \label{tab:train_on_one_eval_on_one}
\end{table*}

\subsubsection{BC ablation}\label{subsec:bc_ablation}

As discussed in Section \ref{sec:result}, the initialization of model training through differential simulation is based on \ttbc model weights.
\begin{table*}[b!]
    \centering
    \caption{Loss ablation study for BC trainings evaluated when controlling \textbf{all agents}. }
    \scriptsize
    \begin{tabular}{l|cccccc} 
        \hline
        Method  &
        \begin{tabular}{@{}c@{}} Col. \\ (\%) \end{tabular}  &
        \begin{tabular}{@{}c@{}} Off. \\ (\%) \end{tabular}  &
        \begin{tabular}{@{}c@{}} ADE \\ (m) \end{tabular}  &
        \begin{tabular}{@{}c@{}} Speed \\ JSD $\times 10^{-2}$ \end{tabular}  &
        \begin{tabular}{@{}c@{}} Acc. \\ JSD $\times 10^{-2}$ \end{tabular}  &
        \begin{tabular}{@{}c@{}} $N_{LC}$ \\ JSD $\times 10^{-2}$ \end{tabular}  \\ \hline
BC wMSE   &     2.76 &       6.36 &   \textbf{1.82} &    \textbf{0.26} &  \textbf{1.84} &  2.16 \\
BC wMSE + Orientation  &  2.71 &  \textbf{3.87} &  3.72 &  3.26 & 8.08 & \textbf{1.29} \\
BC Gaussian-LL  &  2.58 &  5.84 &  2.5 &  1.39 & 16.79 & 2.0 \\
BC GMM-LL & \textbf{2.55} & 5.83 &  2.99 &  2.02 &  20.84 &  1.88 \\

\end{tabular}
    \label{tab:bc_ablation_on_all}
\end{table*}
\begin{table*}[b!]
    \centering
    \caption{Loss ablation study for BC trainings evaluated when controlling \textbf{one agent}. }
    \scriptsize
    \begin{tabular}{l|cccccc} 
        \hline
        Method  &
        \begin{tabular}{@{}c@{}} Col. \\ (\%) \end{tabular}  &
        \begin{tabular}{@{}c@{}} Off. \\ (\%) \end{tabular}  &
        \begin{tabular}{@{}c@{}} ADE \\ (m) \end{tabular}  &
        \begin{tabular}{@{}c@{}} Speed \\ JSD $\times 10^{-2}$ \end{tabular}  &
        \begin{tabular}{@{}c@{}} Acc. \\ JSD $\times 10^{-2}$ \end{tabular}  &
        \begin{tabular}{@{}c@{}} $N_{LC}$ \\ JSD $\times 10^{-2}$ \end{tabular}  \\ \hline
BC wMSE   &     \textbf{0.09} &       0.44 &   \textbf{1.23} &    35.35 &  \textbf{1.38} &  11.63 \\
BC wMSE + Orientation  &  0.13 &  \textbf{0.19} &  2.3 &  6.95 & 9.05 & 10.39 \\
BC Gaussian-LL  &  0.12 &  0.37 &  1.39 &  7.47 & 3.75 & 12.0 \\
BC GMM-LL & \textbf{0.09} & 0.51 &  2.04 &  \textbf{2.37} &  18.65 &  \textbf{10.08} \\

\end{tabular}
    \label{tab:bc_ablation_on_1}
\end{table*}
To establish a better performing model for \ttbc training we perform ablation over different loss functions.
Here, we evaluate the best performing model performance from each method based on their collision and off-road metrics.
The results can be seen in Table \ref{tab:bc_ablation_on_all} and Table \ref{tab:bc_ablation_on_1}.
To leverage the off-road driving performance gains, we initialize the base differential simulation with a \ttbcwMSE method.
\ttmgaildiffsimwmse methods are initialized with \ttbcll weights trained with their respective losses. 

\subsubsection{Feature histograms} \label{subsec: feature histograms} 
\begin{figure*}[h]
    \centering
    
    \subfigure[\tbcgaussll]{\includegraphics[width=0.49\textwidth]{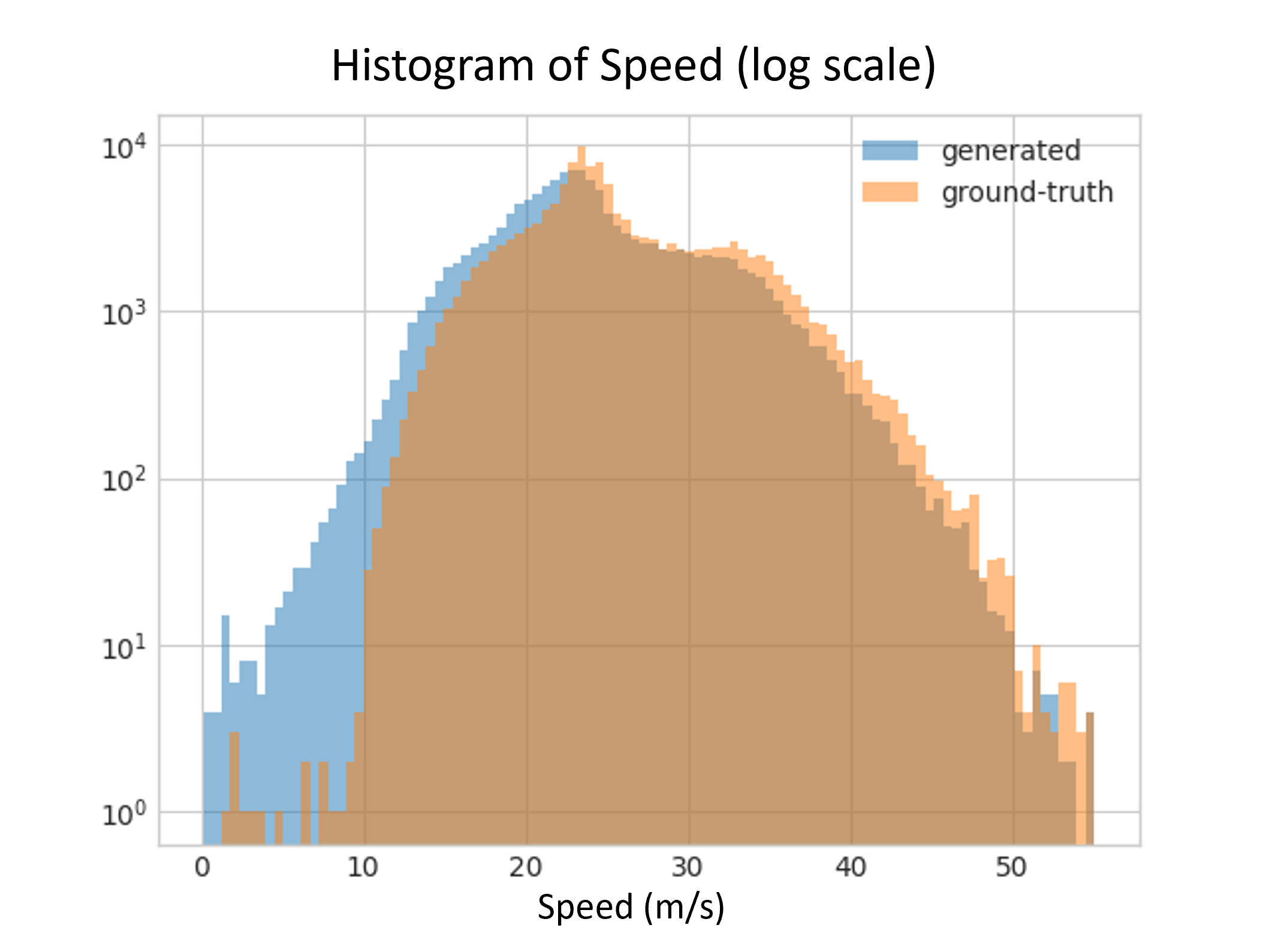}} 
    \subfigure[\tdiffsimwmse]{\includegraphics[width=0.49\textwidth]{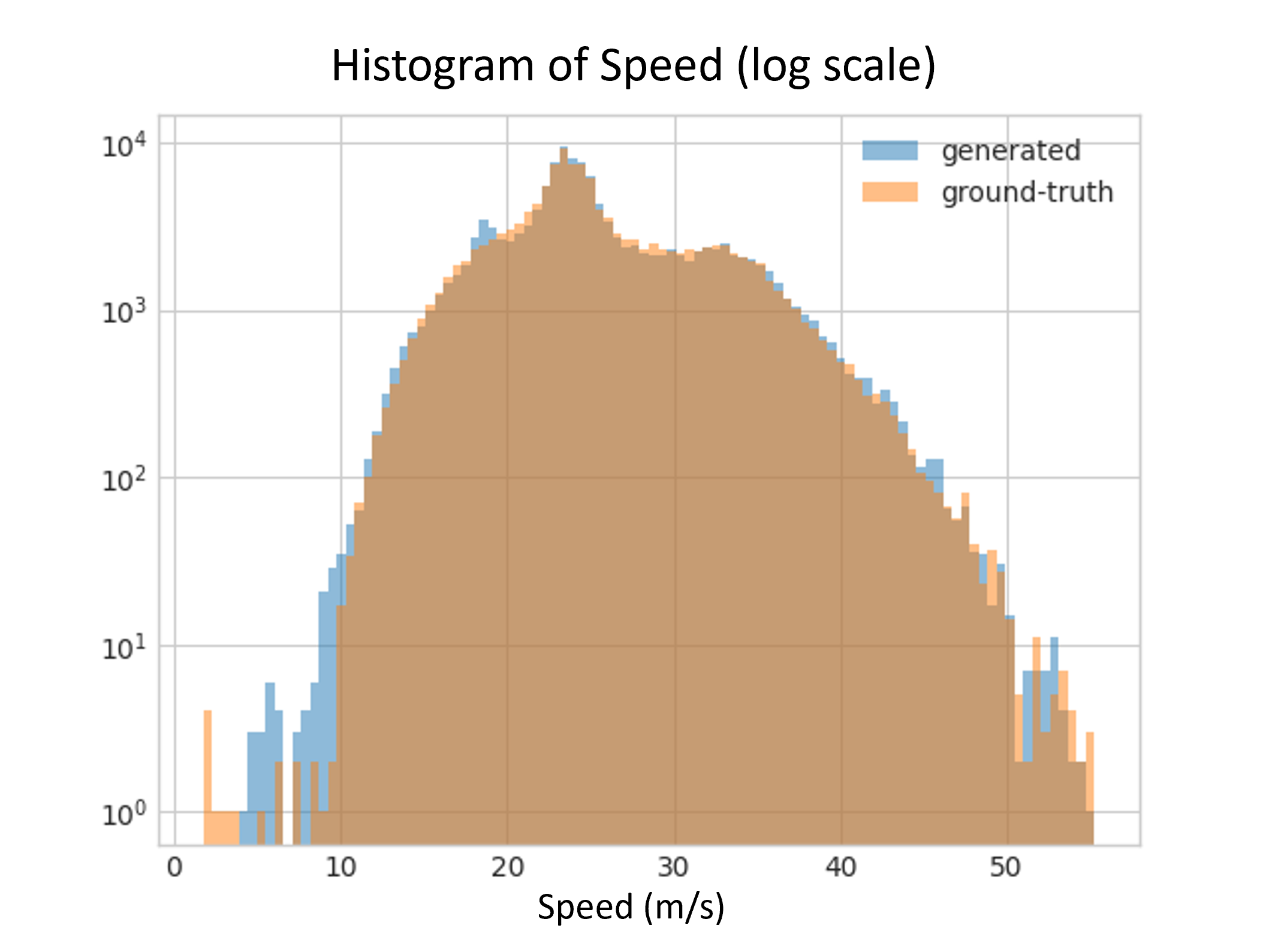}} 
    \subfigure[\tdiffsimwmsecollision]{\includegraphics[width=0.49\textwidth]{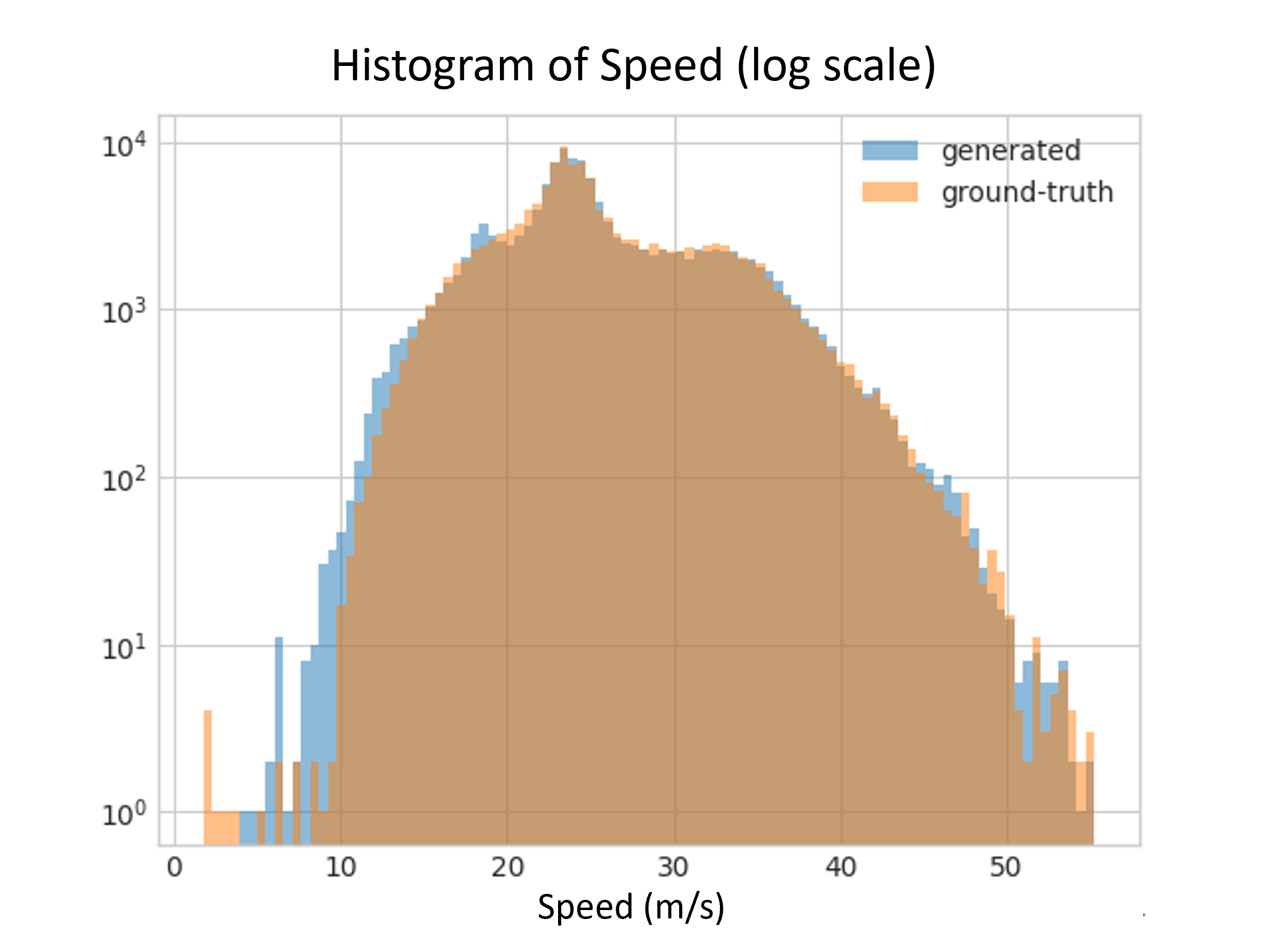}}
    \subfigure[\tmgailbcll (Gaussian)]{\includegraphics[width=0.49\textwidth]{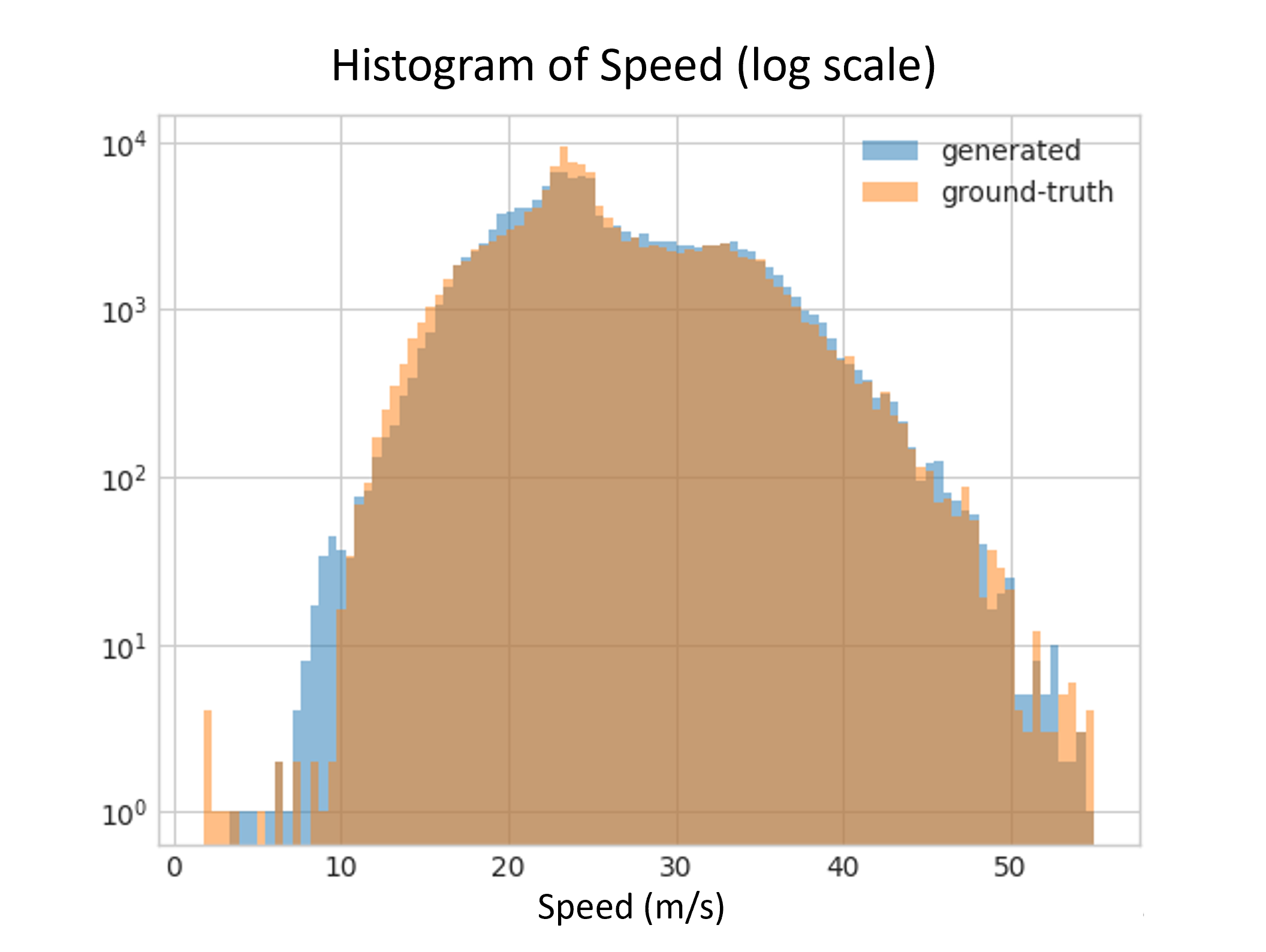}}
    \subfigure[\tmgaildiffsimwmse (Gaussian)]{\includegraphics[width=0.49\textwidth]{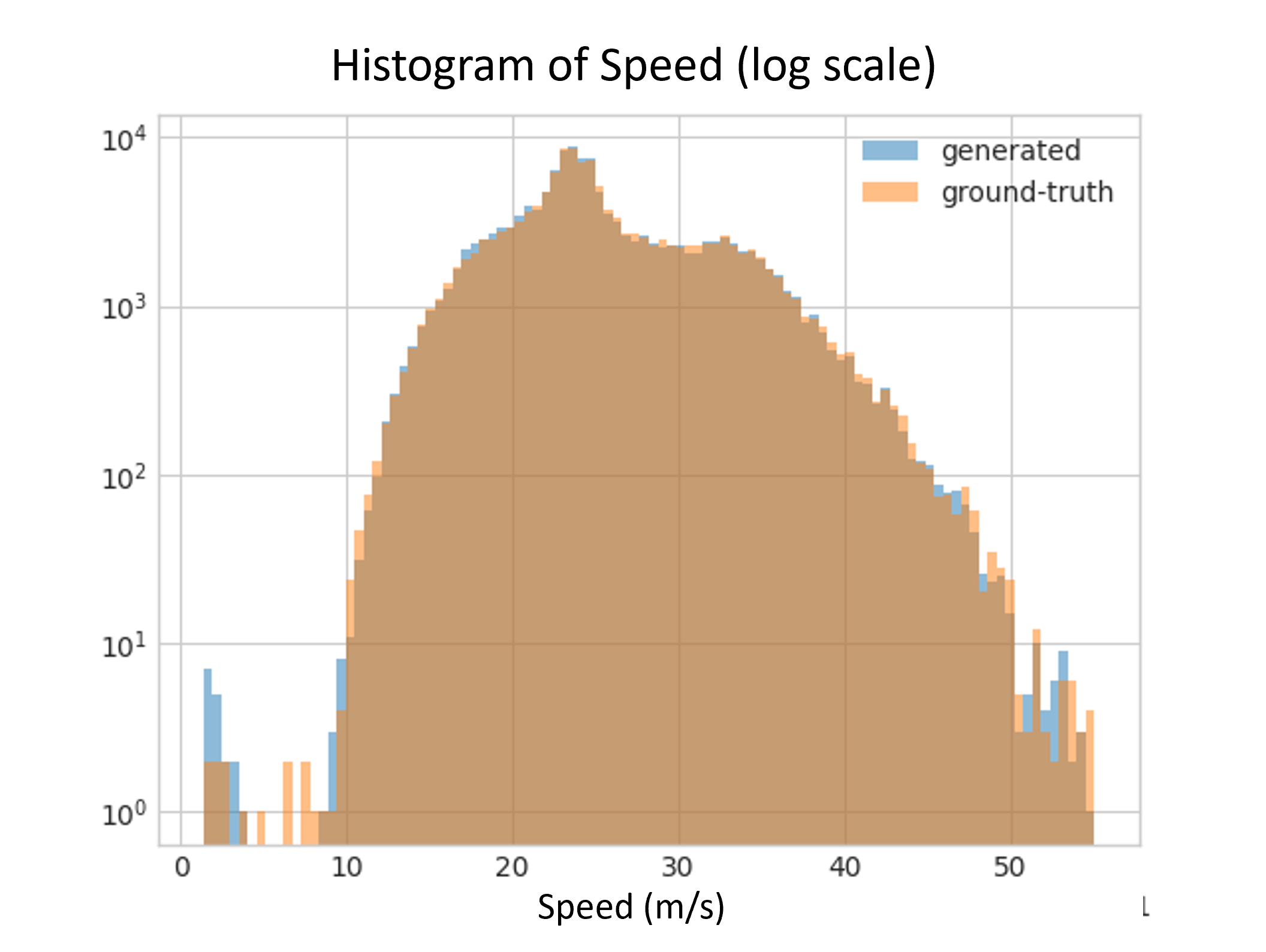}}
    \subfigure[\tmgaildiffsimwmsecollision (Gaussian)]{\includegraphics[width=0.49\textwidth]{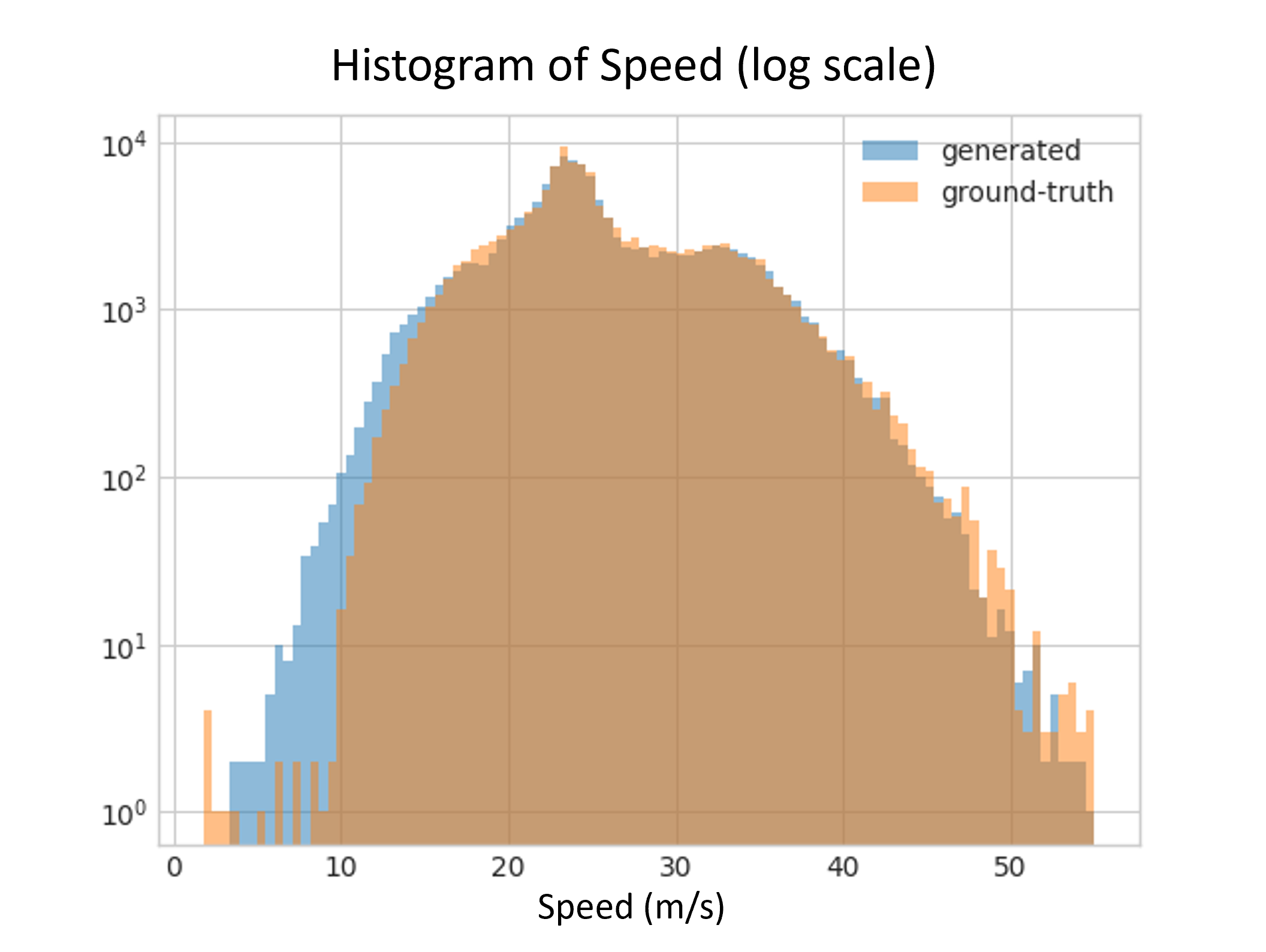}}
    \caption{Speed histograms (log-scale).}
    \label{fig:histograms_realism_speed}
\end{figure*}

\begin{figure*}[h]
    \centering
    
    \subfigure[\tbcgaussll]{\includegraphics[width=0.49\textwidth]{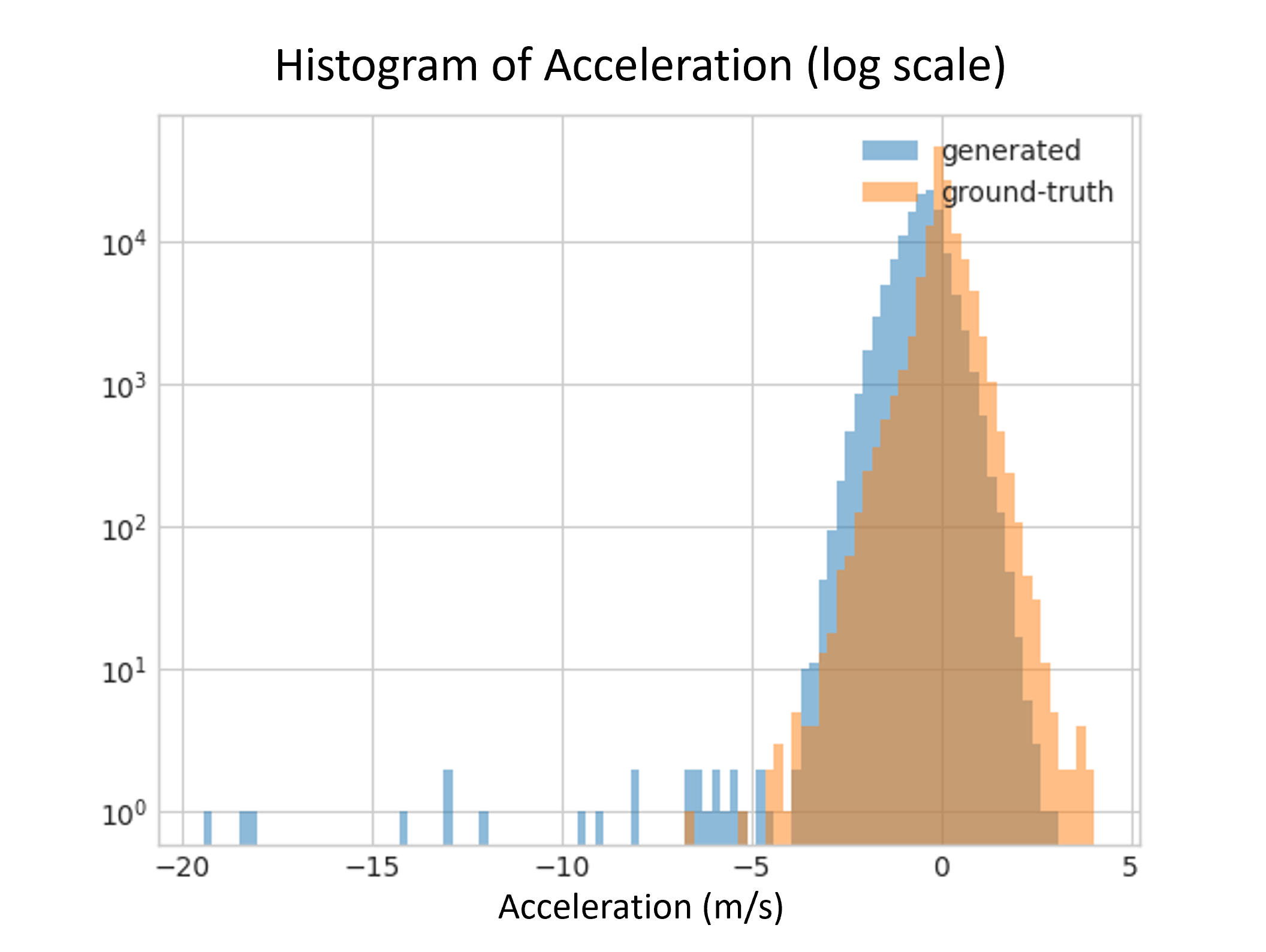}} 
    \subfigure[\tdiffsimwmse]{\includegraphics[width=0.49\textwidth]{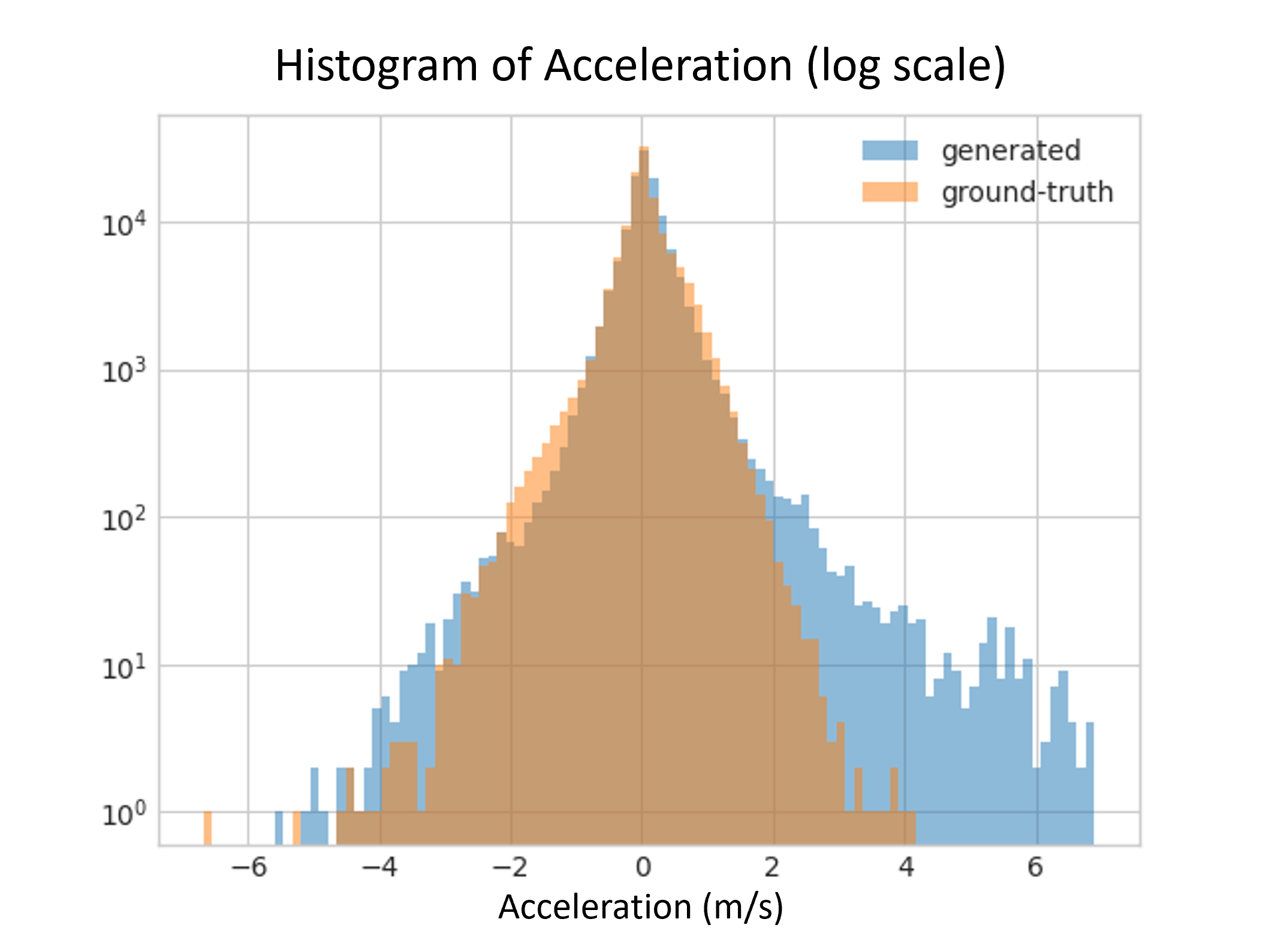}} 
    \subfigure[\tdiffsimwmsecollision]{\includegraphics[width=0.49\textwidth]{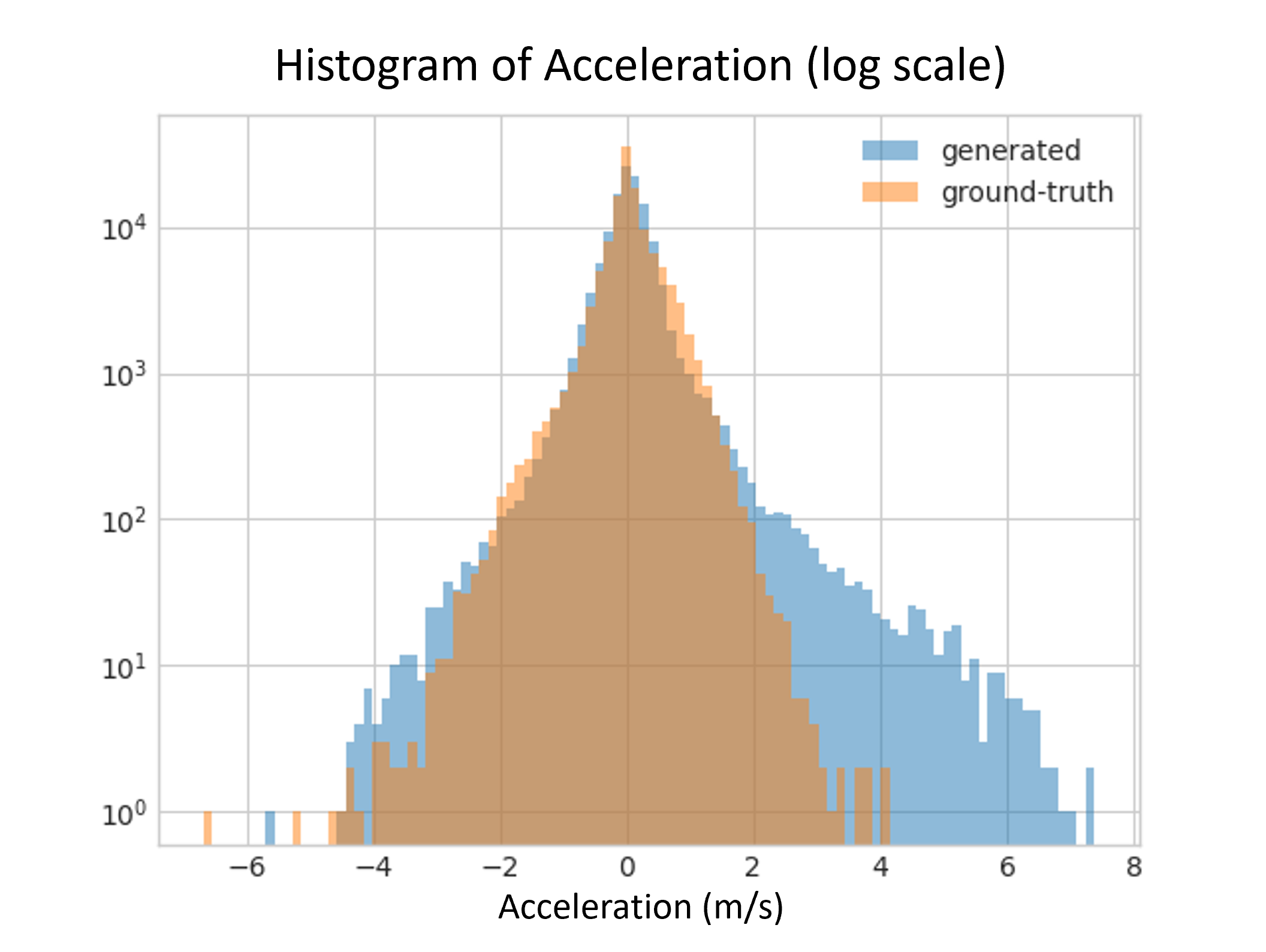}}
    \subfigure[\tmgailbcll (Gaussian)]{\includegraphics[width=0.49\textwidth]{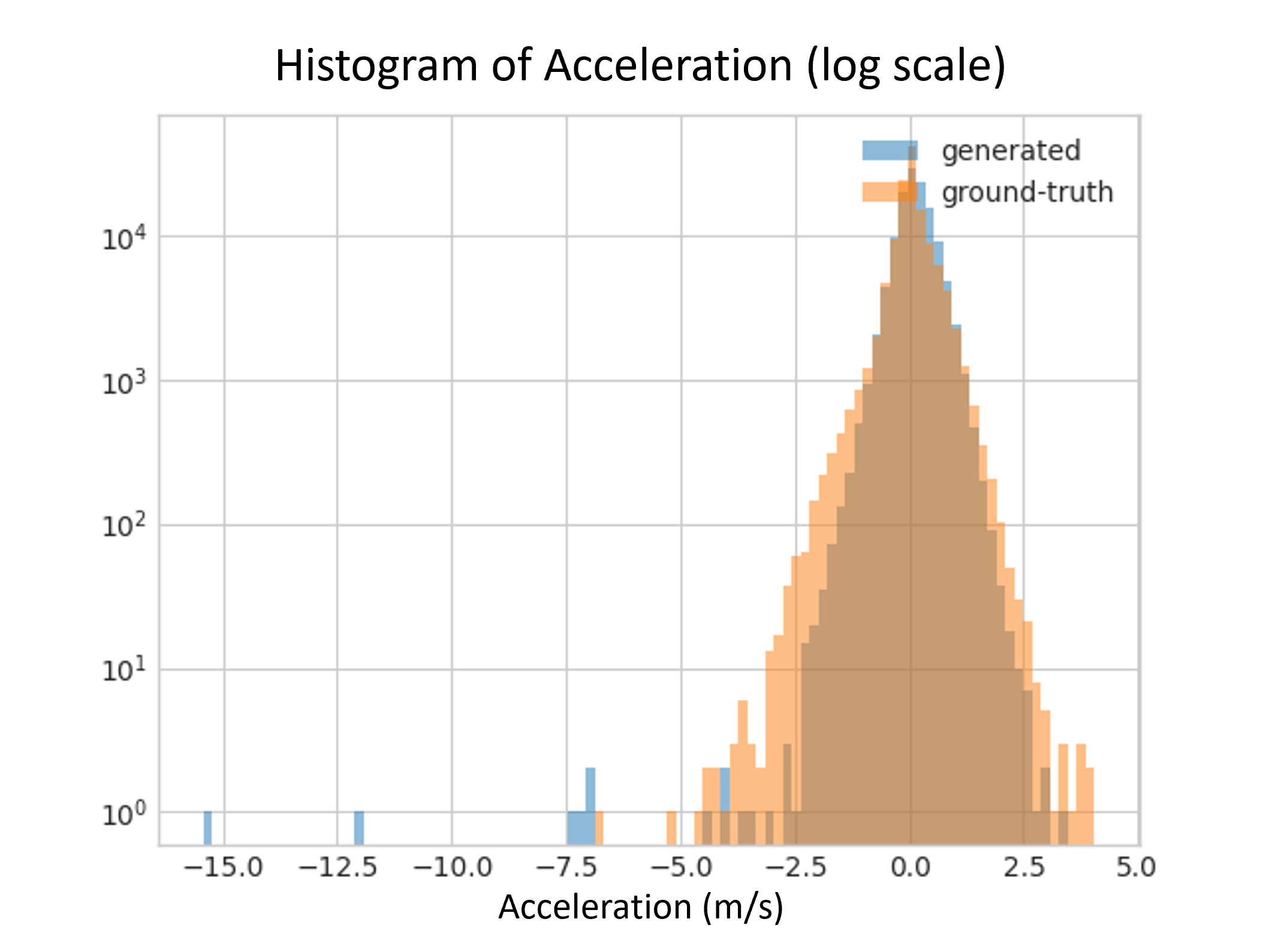}}
    \subfigure[\tmgaildiffsimwmse (Gaussian)]{\includegraphics[width=0.49\textwidth]{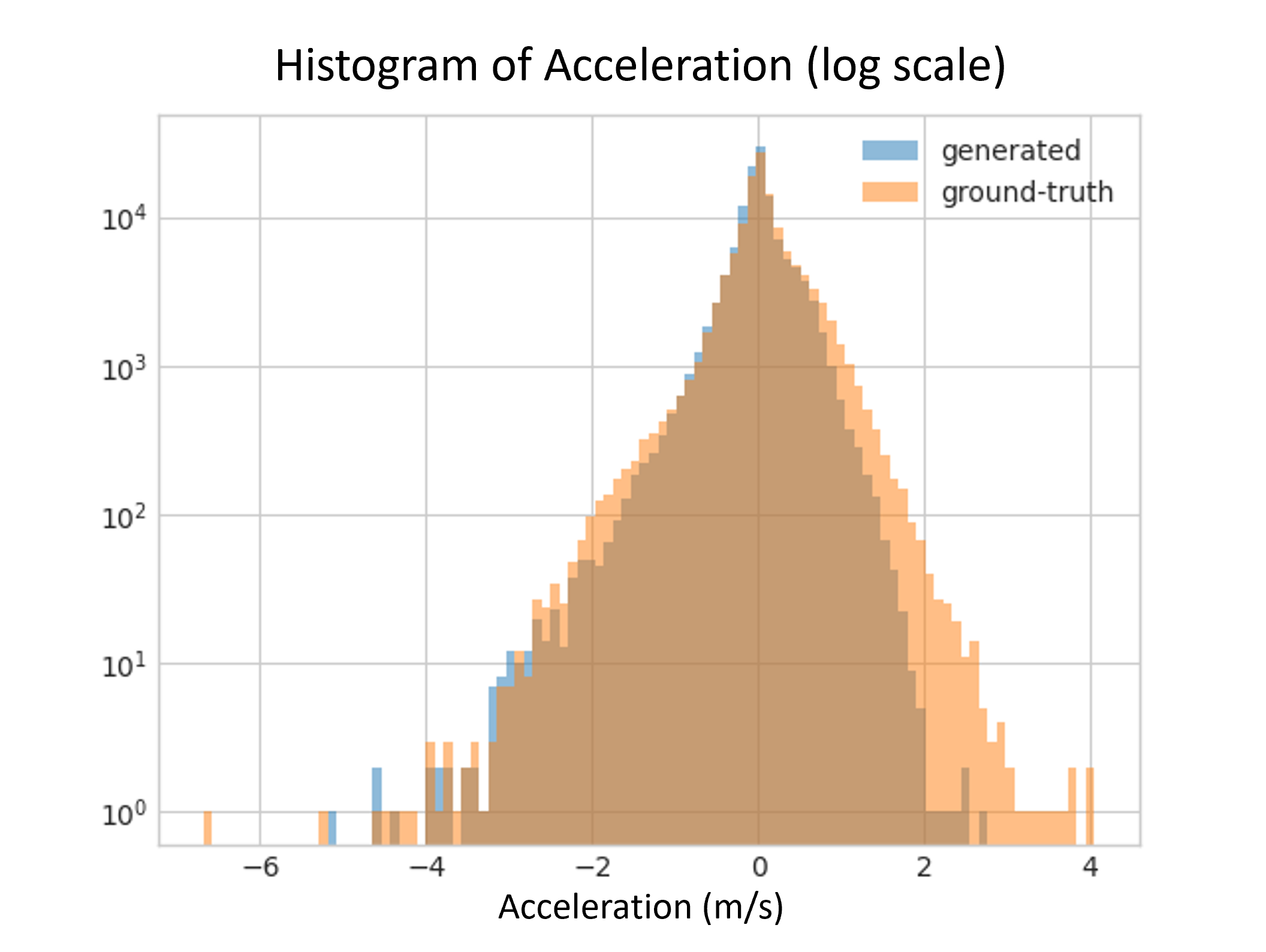}}
    \subfigure[\tmgaildiffsimwmsecollision (Gaussian)]{\includegraphics[width=0.49\textwidth]{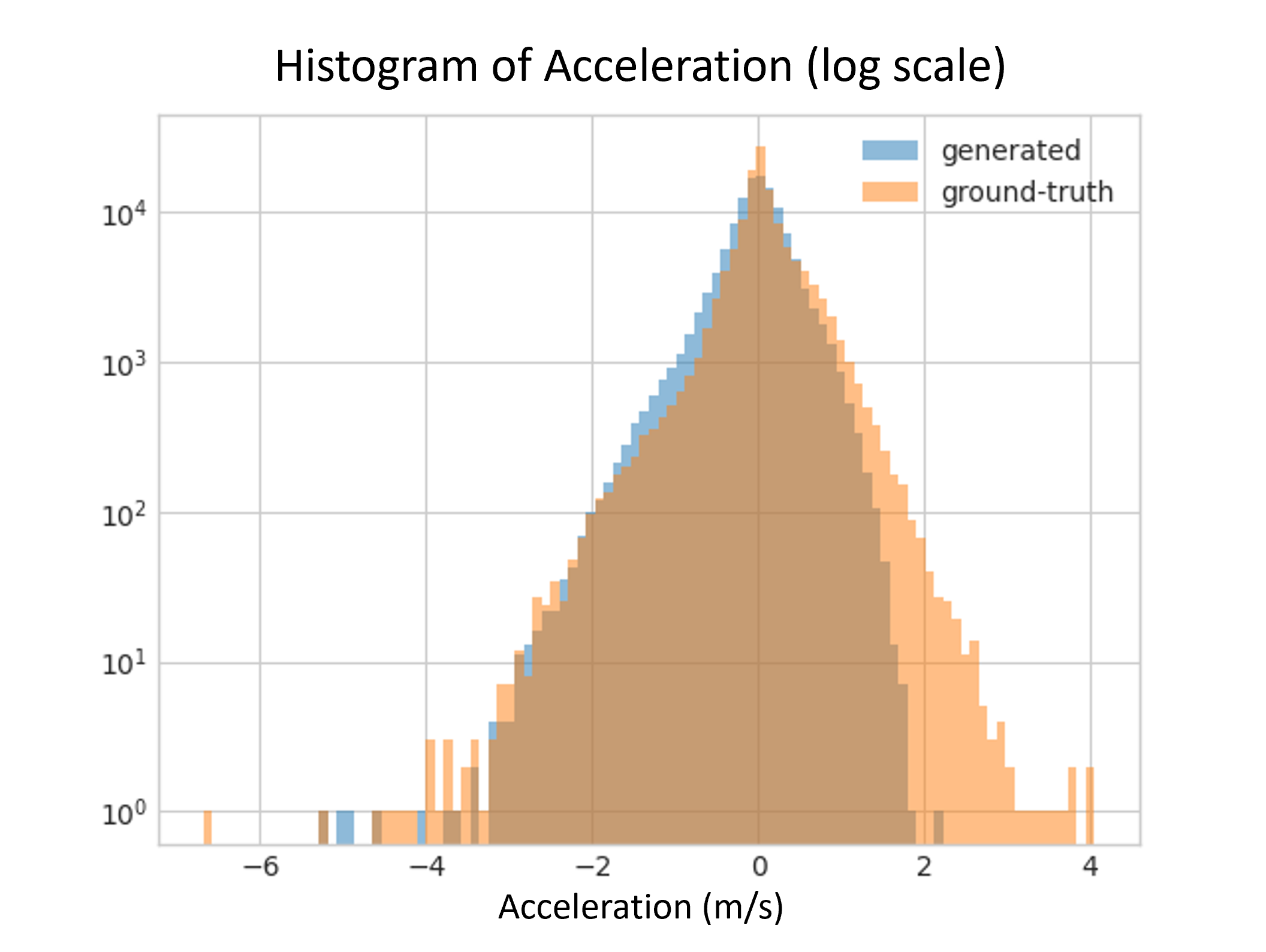}}
    \caption{Acceleration histograms (log-scale).}
    \label{fig:histograms_realism_acc}
\end{figure*}

\begin{figure*}[h]
    \centering
    
    \subfigure[\tbcgaussll]{\includegraphics[width=0.49\textwidth]{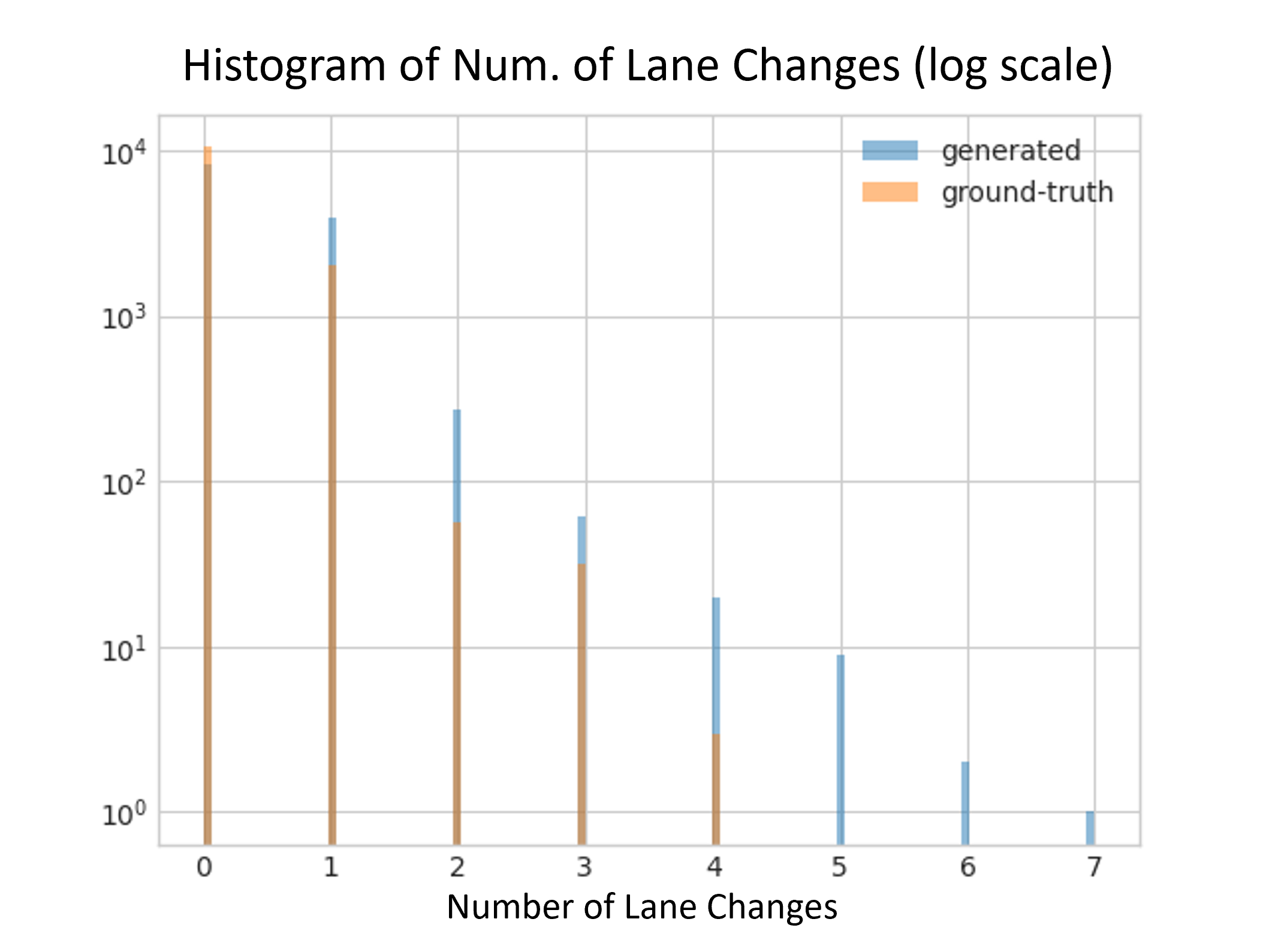}} 
    \subfigure[\tdiffsimwmse]{\includegraphics[width=0.49\textwidth]{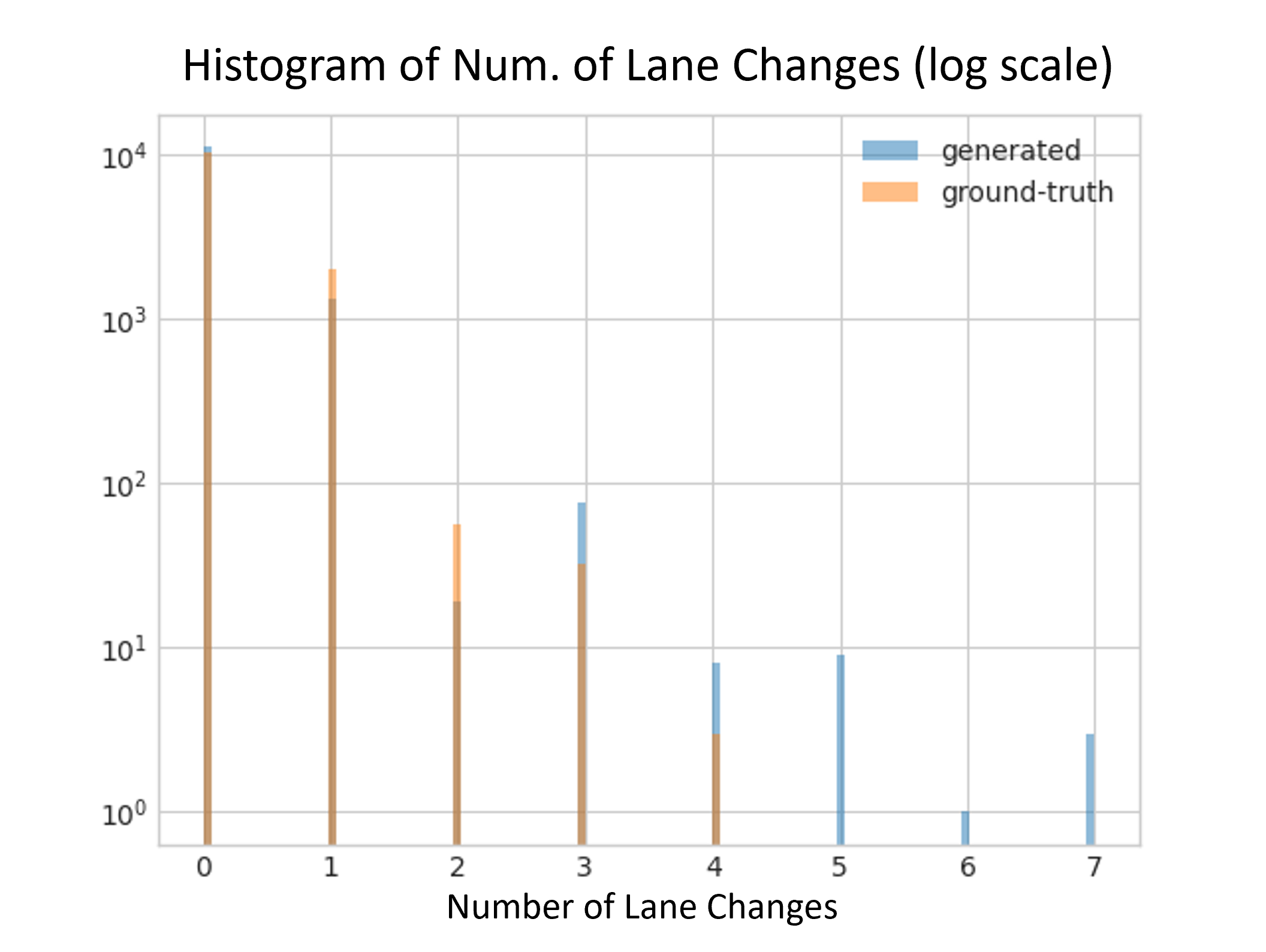}} 
    \subfigure[\tdiffsimwmsecollision]{\includegraphics[width=0.49\textwidth]{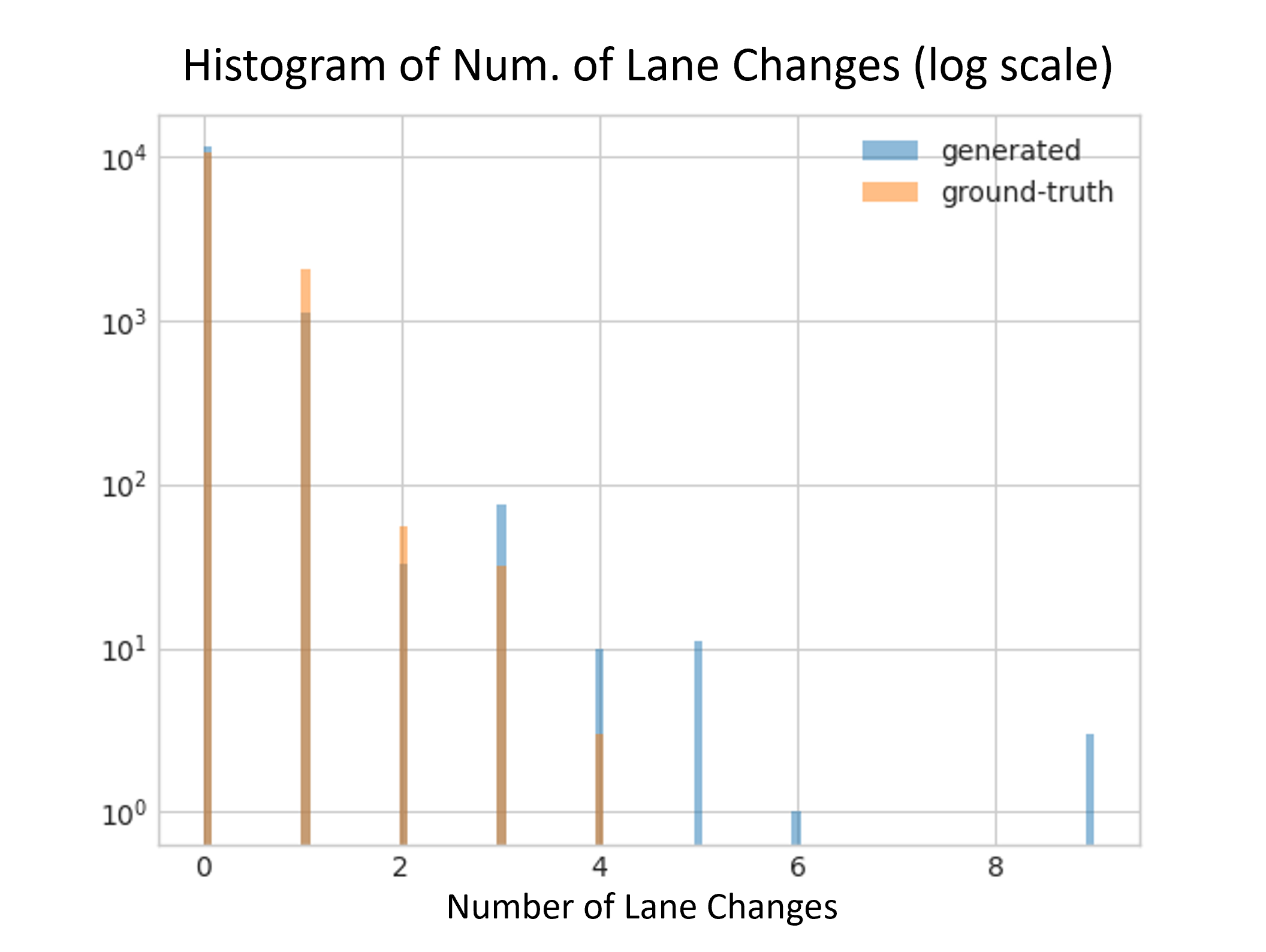}}
    \subfigure[\tmgailbcll (Gaussian)]{\includegraphics[width=0.49\textwidth]{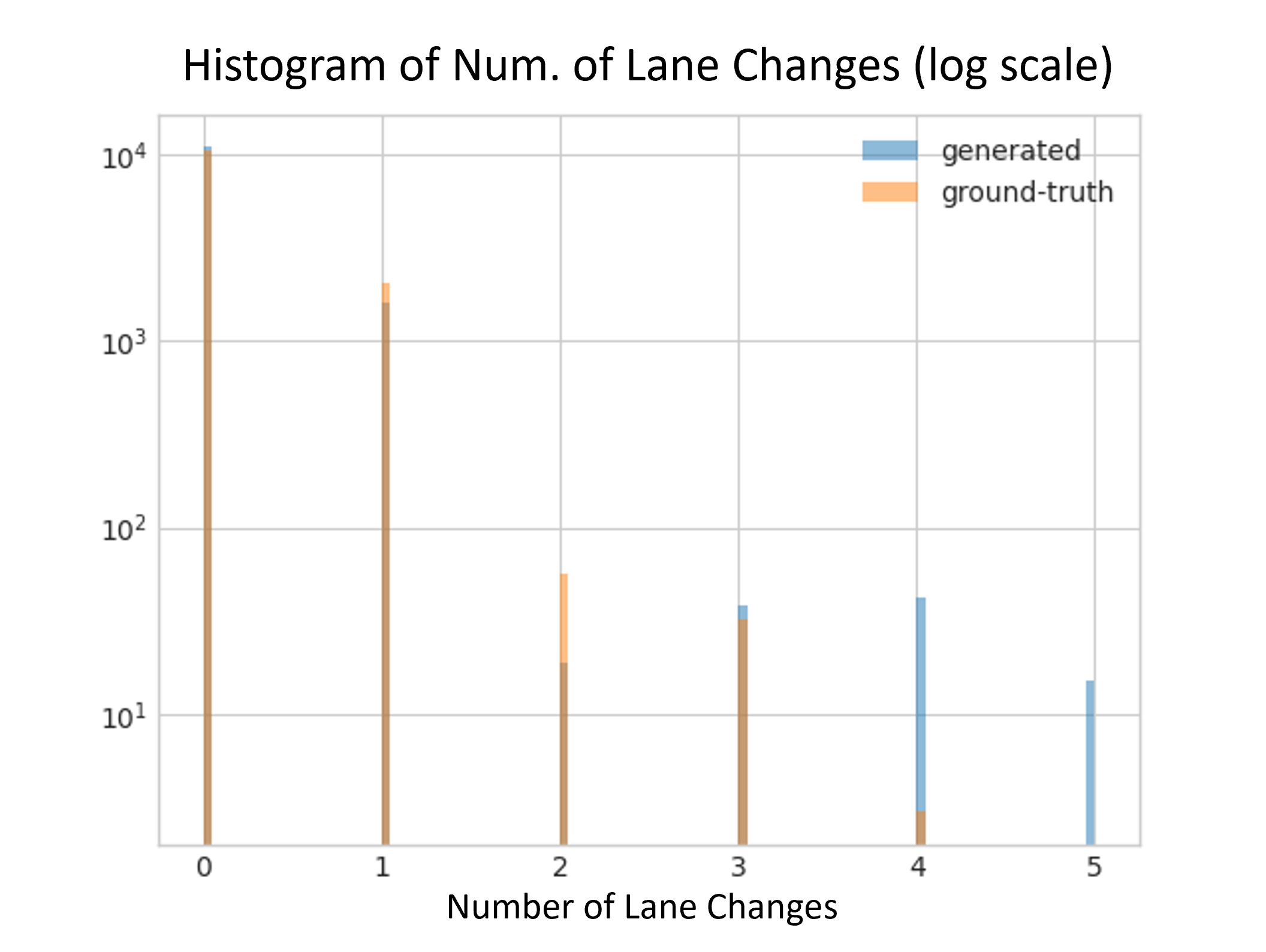}}
    \subfigure[\tmgaildiffsimwmse (Gaussian)]{\includegraphics[width=0.49\textwidth]{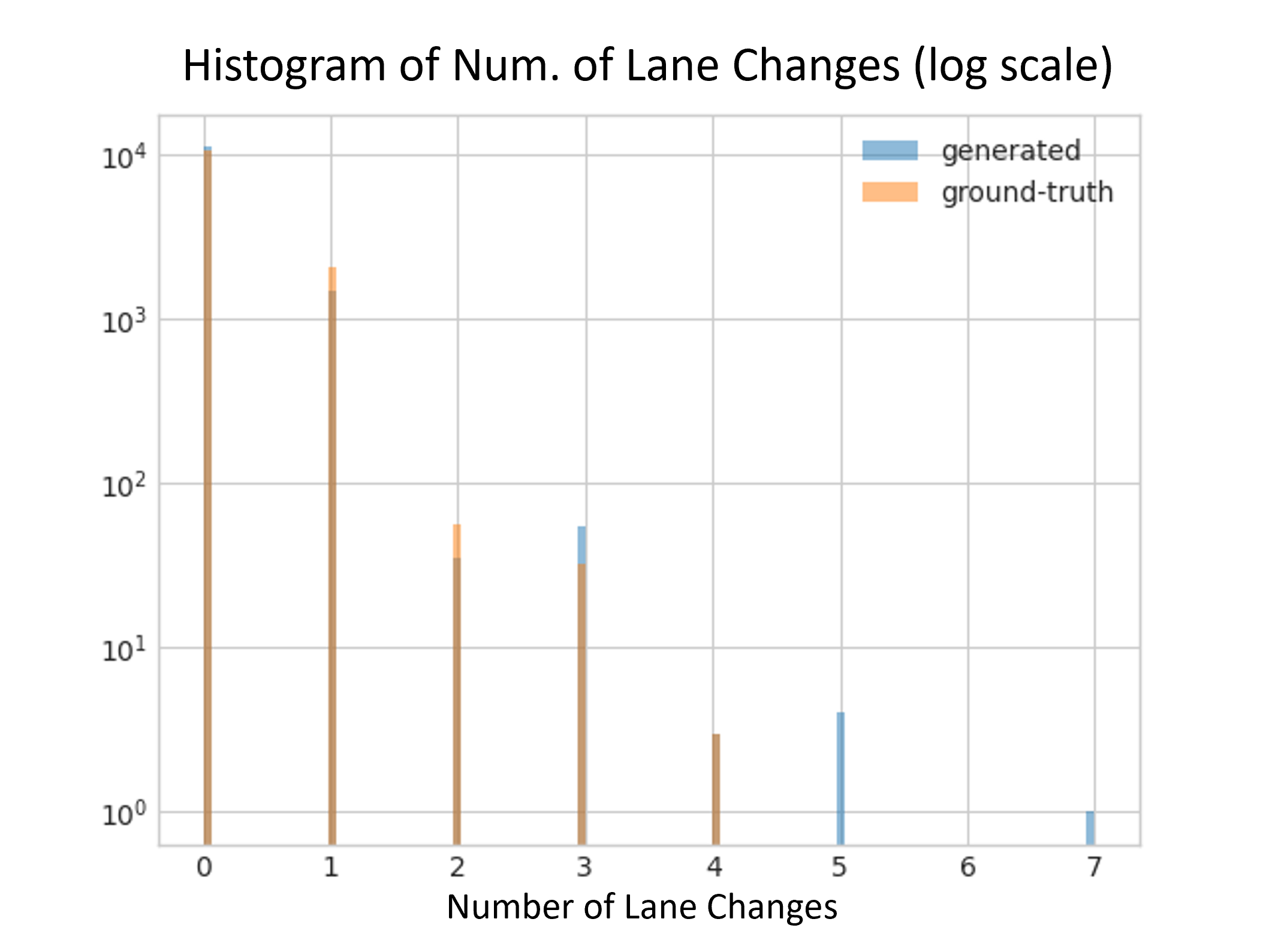}}
    \subfigure[\tmgaildiffsimwmsecollision (Gaussian)]{\includegraphics[width=0.49\textwidth]{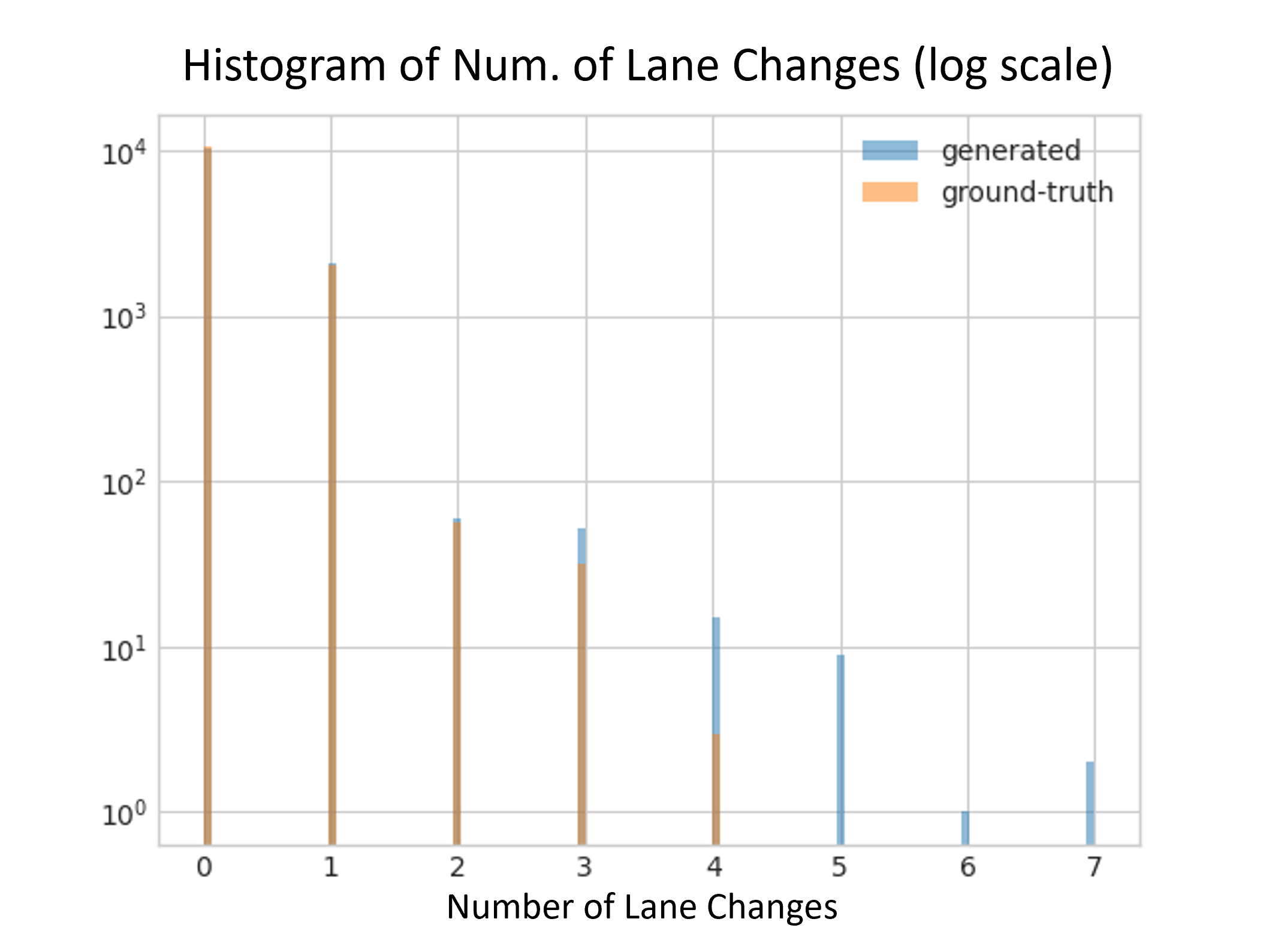}}
    \caption{Number of lane changes histograms (log-scale).}
    \label{fig:histograms_realism_n_ls}
\end{figure*}

We plot histograms of the speed, acceleration and number of lane change features and compare them between ground-truth features and generated ones. In Figure \ref{fig:histograms_realism_speed}, the speed histogram is shown; in Figure \ref{fig:histograms_realism_acc} the acceleration histogram is shown; and in Figure \ref{fig:histograms_realism_n_ls} the histogram over the number of lane changes. We show the ground-truth histogram in orange and the generated histograms in blue, and we render histograms for the generated features over several methods. For each method, we use the run with median value with respect to the corresponding JSD value (out of five runs) for histogram plotting. All histograms are shown in log-scale.

\end{document}